\documentclass[authoryear,preprint,12pt]{elsarticle}
\usepackage[pdfborder={0 0 0},colorlinks]{hyperref} %
\usepackage{amsthm,amsmath,amsfonts,latexsym,amssymb}
\usepackage{graphics,fancyhdr,graphicx}
\usepackage[ruled,linesnumbered]{algorithm2e}
\usepackage[margin=1in]{geometry}
\usepackage{multirow,booktabs}
\usepackage[table]{xcolor}
\usepackage{listings}
\usepackage{tabularx}
\usepackage{placeins}
\usepackage{comment}
\usepackage{lineno}
\usepackage{lscape}
\usepackage{bm}
\usepackage[labelfont=bf]{caption}
\usepackage{float}
\usepackage{soul}
\usepackage{setspace} 
\hypersetup{
    pdfauthor  = author
    urlcolor   = black,
    citecolor  = black,
}
\usepackage{textgreek}
\usepackage[utf8]{inputenc}
\usepackage[T1]{fontenc}
\biboptions{numbers,sort&compress}
\usepackage{lineno}
\usepackage{array,makecell,threeparttable}

\emergencystretch=\maxdimen
\newcommand{\beginsupplement}{
    \setcounter{table}{0}
    \renewcommand{\thetable}{S\arabic{table}}
    \setcounter{figure}{0}
    \renewcommand{\thefigure}{S\arabic{figure}}
    \renewcommand{\thesection}{S}
}

\begin{document}

\begin{frontmatter}

\title{Importance of localized dilatation and distensibility in identifying determinants of thoracic aortic aneurysm with neural operators}

\author[1]{David S. Li}
\author[2]{Somdatta Goswami}
\author[3]{Qianying Cao}
\author[4]{Vivek Oommen}
\author[5]{\\Roland Assi}
\author[1,6]{Jay D. Humphrey\corref{correspondingauthor}}
\author[3]{George E. Karniadakis}

\cortext[correspondingauthor]{Corresponding author: jay.humphrey@yale.edu}

\address[1]{Department of Biomedical Engineering, Yale University, 55 Prospect St, New Haven, CT 06511, USA}
\address[2]{Department of Civil and Systems Engineering, Johns Hopkins University, 3400 N Charles St, Baltimore, MD 21218, USA}
\address[3]{Division of Applied Mathematics, Brown University, 170 Hope St, Providence, RI 02906, USA}
\address[4]{School of Engineering, Brown University, 184 Hope St, Providence, RI 02912, USA}
\address[5]{Department of Surgery, Yale School of Medicine, 800 Howard Ave, New Haven, CT 06519, USA}
\address[6]{Vascular Biology and Therapeutics Program, Yale School of Medicine, 10 Amistad St, New Haven, CT 06520, USA}

\begin{abstract}
\noindent Thoracic aortic aneurysms (TAAs) stem from diverse mechanical and mechanobiological disruptions to the aortic wall that can also increase the risk of dissection or rupture. There is increasing evidence that dysfunctions along the aortic mechanotransduction axis, including reduced integrity of elastic fibers and loss of cell-matrix connections, are particularly capable of causing thoracic aortopathy. Because different insults can produce distinct mechanical vulnerabilities, there is a pressing need to identify interacting factors that drive progression. In this work, we employ a finite element framework to generate synthetic TAAs arising from hundreds of heterogeneous insults that span a range of compromised elastic fiber integrity and cellular mechanosensing. From these simulations, we construct localized dilatation and distensibility maps throughout the aortic domain to serve as training data for neural network models to predict the initiating combined insult. Several candidate architectures (Deep Operator Networks, UNets, and Laplace Neural Operators) and input data formats are compared to establish a standard for handling future subject-specific information. We further quantify the predictive capability when networks are trained on geometric (dilatation) information alone, which mimics current clinical guidelines, versus training on both geometric and mechanical (distensibility) information. We show that prediction errors based on dilatation data are significantly higher than those based on dilatation and distensibility across all networks considered, highlighting the benefit of obtaining local distensibility measures in TAA assessment. Additionally, we identify UNet as the best-performing architecture across all training data formats. These findings demonstrate the importance of obtaining full-field measurements of both dilatation and distensibility in the aneurysmal aorta to identify the mechanobiological insults that drive disease progression, which will advance personalized treatment strategies that target the underlying pathologic mechanisms.
\end{abstract}




\end{frontmatter}



\section*{Author summary}
We investigated causes of thoracic aortic aneurysms (TAAs), which are local enlargements of the aorta that can lead to life-threatening rupture. TAAs result from a combination of structural and cellular disruptions in the aortic wall, but need not pose the same risk even if appearing similar in size. We created computer models of aneurysms caused by different combinations of known risk factors, including tissue damage and cellular dysfunction. Using these simulations, we trained multiple neural network models with information on TAA shape (dilatation) and mechanics (distensibility) to see if they could recover the original causes. We found that predictions based on both dilatation and distensibility were significantly better than using dilatation alone, which represents the current clinical standard. Among the models tested, we also identified a best performing architecture, UNet, for these applications. These results suggest that measuring the full shape and mechanics of the aneurysmal aorta could lead to improved personalized diagnoses and treatment for TAAs.

\section*{Introduction}
Treatment planning for individuals with thoracic aortic aneurysm (TAA), a localized dilatation resulting from underlying microstructural damage, continues to be based primarily on aortic size and growth rate \cite{Senser2021}, with maximum size being the \emph{de facto} predictor. Nevertheless, life-threatening aortic events occur below established thresholds \cite{Kim2016, Mansour2018}. Importantly, there is increasing evidence that dysfunction can occur at several points along the mechanotransduction axis \cite{Humphrey2014, Humphrey2015, Karimi2016, Yamashiro2020, Creamer2021} that informs aortic cells responsible for maintaining healthy mural structure and function. These may include reduced or dysfunctional fibrillin-1 that stabilizes elastic fibers in the aortic wall \cite{LopezGuimet2017, Cavinato2021}, loss of cell-matrix connections (microfibrils and integrin binding sites) required for accurate assessment of wall stress and assembly of extracellular matrix \cite{Li2003, Turlo2012}, disrupted deposition or organization of fibrillar collagens \cite{Weiss2023}, aberrant transforming growth factor-beta signaling \cite{Ramirez2018}, and altered cellular contractility \cite{Chung2007, Milewicz2017}. Many of these effects have been shown to result in aneurysmal dilatation; yet, because different insults may lead to lesions of similar size but vastly different mechanical strength and vulnerability, there is a critical need to understand the underlying biomechanical and mechanobiological mechanisms that drive disease progression in a patient-specific manner. Moreover, since mechanobiological function (and its compromise) cannot be directly evaluated \emph{in vivo}, such insight must be gained using often-limited, minimally-invasive clinical information following diagnosis.

Toward this end, we recently assessed the performance of neural network models in identifying factors that could contribute to TAA in a synthetic dataset resulting from finite element (FE) simulations, using metrics that can be derived simply from three-dimensional medical image reconstructions \cite{Goswami2022}. Briefly, a simulation platform for aortic growth and remodeling (G\&R) \cite{Latorre2020} was used to generate aneurysms from randomly distributed, localized insults to structural constituents and mechanobiological mechanisms in a healthy aortic model. Local dilatation (normalized inner radius) and distensibility (normalized difference in radius between systole and diastole) fields from the resulting TAAs were used as training data for a deep operator network (DeepONet) \cite{Lu2021} in the form of two-dimensional maps, which revealed the efficacy of a convolutional neural network (CNN)-based DeepONet architecture in predicting the associated initiating insult profile for a given dilatation-distensibility pair with a high degree of accuracy, thereby establishing the feasibility of evaluating mechanobiological compromise using image-based information.

Yet, experimental observations from disease models usually cannot be captured with a computational framework without consideration of multi-contributor insults \cite{Weiss2021,Li2023}. For instance, recent work elucidating mechanobiological drivers of TAA in the genetic condition Marfan syndrome \cite{Pereira1999, Judge2004} identified several interconnected effects in its natural history, including both structural (e.g., elastic and collagen fibers) and biological (e.g., cellular mechanosensing and mechanoregulation of matrix) impairments in severe disease \cite{Cavinato2021, Li2023}. It is clear that synthetic data for training surrogate models must similarly consist of combined effects to yield clinically useful predictions. To more closely represent the scope of \emph{in vivo} pathologies leading to TAA, we now build on this established framework by introducing additional factors to our synthetic data generation pipeline, including parameterization of the model to ascending aortic biomechanical properties of a mouse model of TAA (prone to dissection and rupture), as well as, importantly, focusing on aneurysms resulting from multiple superimposed initiators.

Furthermore, looking beyond the DeepONet framework, we wish to evaluate a range of candidate neural operators varying in type and architecture (including UNets and Laplace Neural Operators) that may be best suited for these applications. We compare the performance of several architectures in identifying these multi-contributor insult profiles from maps of local dilatation and distensibility. Importantly, to bracket the range of predictive capability in the lower limit of available clinical information, each network is also trained with knowledge of dilatation alone. We show that both dilatation and distensibility information are necessary for accurate estimation of combined insult profiles, especially for TAAs that exhibit similar maximum dilatations. When assessing the predictive accuracy of these models when trained on both dilatation and distensibility data versus dilation data only, we find that the prediction accuracy is significantly lower when relying solely on dilatation across all networks. Additionally, models based on convolutional neural networks, particularly UNets, exhibit the best performance in determining mechanobiological insult magnitude and distribution, serving as a promising tool for predicting TAA determinants and gaining insight into patient risk.

\section*{Materials and methods}
\subsection*{Generation of synthetic data}
Computational studies on TAA have investigated the propensity of multiple localized perturbations in aortic mechanics and mechanobiology to initiate and propagate dilatations in the aorta, including loss of elastic fiber integrity \cite{Wilson2012, Weiss2021, Laubrie2022}, reduced collagen cross-linking \cite{Wilson2013, Watton2004}, dysfunctional mechanosensing \cite{Humphrey2014}, and impaired mechanoregulation of collagen fibers \cite{Irons2022}. While these effects have largely been studied individually, with some recent effort devoted toward combining them to capture experimental observations \cite{Mousavi2021, Li2023}, neural network models have yet to be trained on data comprised of these interacting contributors. We leverage our existing pipeline described elsewhere \cite{Humphrey2021, Latorre2018a, Latorre2020, Murtada2021}, built upon the notion of mechanobiological homeostasis and its loss in the aorta, with key extensions detailed below. Further information may be found in \hyperref[sec:SI]{S1 Supporting Information}.

\subsubsection*{Contributing insults}
We focus on two primary classes of contributing factors to TAA \cite{Goswami2022}, namely reduced integrity of elastic fibers in the aortic wall and compromised mechanosensing of intramural cells responsible for modulating the mechano-response of the aorta. Both effects have been shown to play critical roles in aneurysmal progression, leading to localized dilatation, abberant elastic energy storage, and increased circumferential stiffness \cite{Bellini2017, Latorre2020, Li2023}. As in previous work, insults to elastic fibers take the form of prescribed reductions in the material parameter $c^e$ representing the stiffness of the elastin-dominated extracellular matrix (Table A in \hyperref[sec:SI]{S1 Supporting Information}), particularly impacting the energy storage capability of the aorta required for the windkessel effect in normal function. Mechanosensing insults are characterized by scaling the deviation in the intramural stress $\Delta \sigma = (\sigma/\sigma_o) - 1$ that is ``sensed'' by aortic cells, where $\sigma$ is a scalar measure and $\sigma_o$ is the homeostatic ``set-point'' that the cells work to maintain by modulating cell and matrix turnover in the aortic wall. Specifically, a coefficient $\delta \in [0,1]$ is introduced in the expression to give $\Delta \sigma = ((1-\delta)\sigma/\sigma_o) - 1$ (with $\delta = 0$ being perfect mechanosensing in the normal aorta), capturing impairment of the ability of the aorta to effectively modify the surrounding matrix constituents in response to elevated intramural stress \cite{Turlo2012, Latorre2020, Li2023}.

\begin{figure}[!t]
\centering
\includegraphics[width=\textwidth]{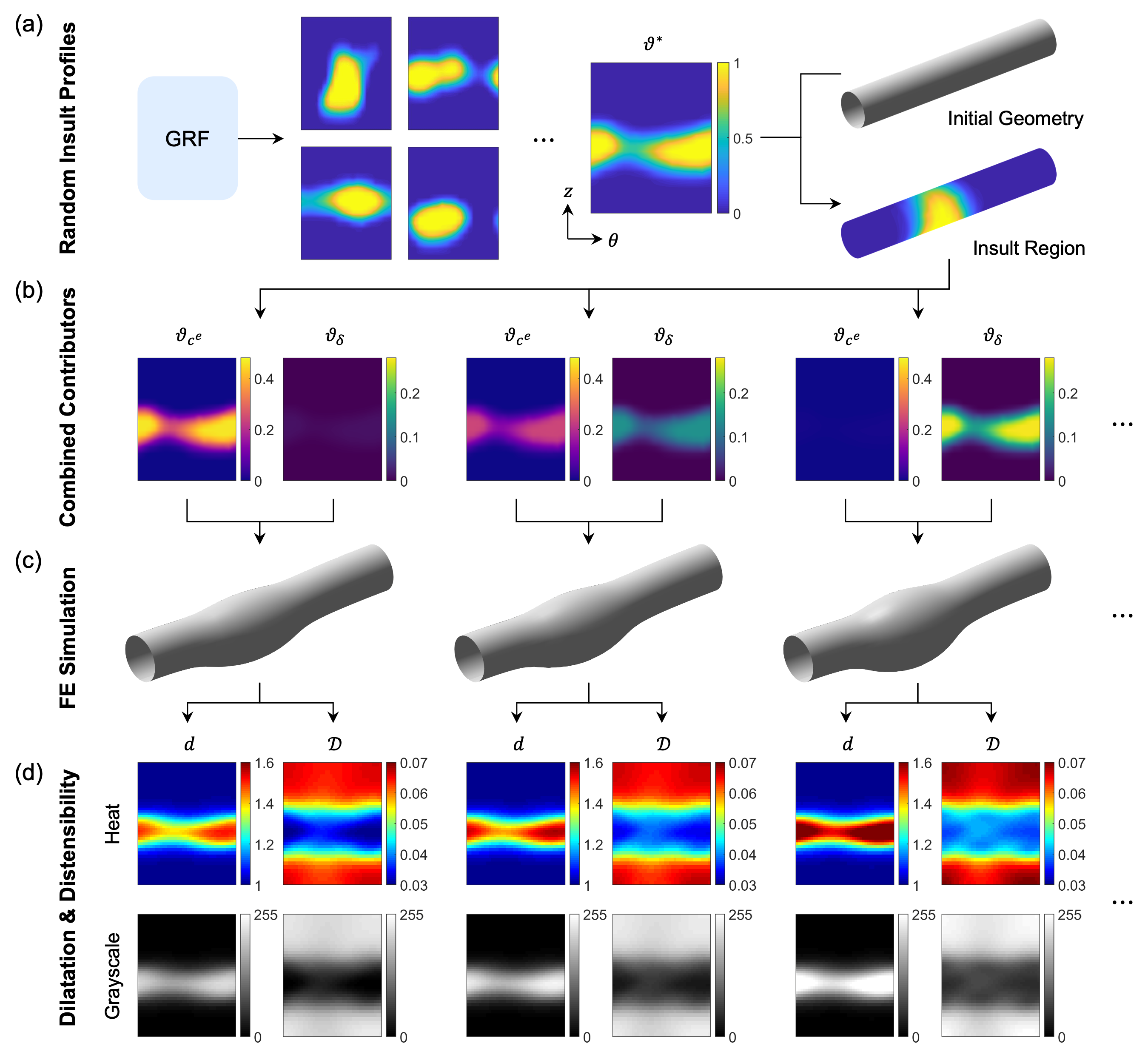}
\caption{{\bf Synthetic data generation pipeline.} (a) Insult distributions along the circumferential ($\theta$) and axial ($z$) directions are randomly generated with Gaussian Random Field (GRF) methods to define normalized insult profiles ($\vartheta^\ast(\theta,z) \in [0,1]$) that are mapped to the initial loaded aortic geometry to define the insult region (see \hyperref[sec:SI]{S1 Supporting Information} for further details). (b) For each profile, $\vartheta^{\ast}$ delineates multiple cases of mechanobiological insults defined by combinations of compromised integrity of elastic fibers ($\vartheta_{c^e}(\theta,z) \in [0,0.48]$) and dysfunctional mechanosensing ($\vartheta_{\delta}(\theta,z) \in [0,0.28]$) as inputs to (c) the nonlinear FE simulation to compute the long-term evolved state of the TAA. (d) Maps for dilatation ($d$) and distensibility ($\mathcal{D}$) are obtained from the final geometry under multiple \textit{in vivo} loading conditions (diastolic and systolic pressures), either processed as heat maps or converted to 8-bit grayscale maps, to serve as training data for the neural networks. $\vartheta_{c^e} + \vartheta_{\delta}$ combinations are constrained to produce closely matched maximum dilatations ($d_{max} \approx 1.5$) for each case to focus attention on detecting the underlying pathologic mechanism at the time when a dilatation first reaches aneurysmal status.}
\label{fig:data-generation}
\end{figure}

To generate synthetic training data, we employed a well-established computationally efficient FE model for determining the long-term, mechanobiologically equilibrated evolution of TAAs \cite{Latorre2020}, with baseline parameters estimated to reproduce the \textit{in vivo} behavior of a non-dilated ascending aorta of a mouse model of Marfan syndrome under normotensive conditions \cite{Li2023}. We used our recently developed method for mapping spatially heterogeneous perturbations generated via Gaussian Random Fields to produce randomly distributed insult profiles $\vartheta^\ast(\theta,z) \in [0,1]$ (\autoref{fig:data-generation}a), which indicate the local normalized severity of mechanobiological compromise throughout the initial aortic domain. Each profile defines spatially varying multi-contributor insults by superimposing defects in elastic fiber integrity $\vartheta_{c^e}(\theta,z)$ and mechanosensing $\vartheta_\delta(\theta,z)$ (\autoref{fig:data-generation}b). Each contributor pair was provided as input to the FE TAA simulation (\autoref{fig:data-generation}c) to compute the steady-state evolved geometry, at which point the G\&R evolution was halted, and we simulated changes in luminal pressure over a cardiac cycle. Dilatation $d$ was defined as the local inner radius $r$ from the aortic centerline normalized by the average inner radius at the vessel ends, and distensibility $\mathcal{D} = (r_S - r_D)/r_D$ was defined as the normalized change in inner radius between systolic ($S$) and diastolic ($D$) loading (\autoref{fig:data-generation}d, cf. \cite{Goswami2022}), both of which can be computed from a cardiac gated medical image. The magnitudes of the insult contributors $\vartheta_{c^e,\delta}$ were constrained to produce maximum dilatations of approximately 1.5 (defined as aneurysmal), with closely matched dilatation values across all insult combinations to focus attention on detecting the underlying pathologic mechanism at the time that a dilatation first reaches aneurysmal status, a critical time in clinical treatment planning (Fig A in \hyperref[sec:SI]{S1 Supporting Information} displays the differential effects of a wider range of combinations of $\vartheta_{c^e,\delta}$ on dilatation and distensibility). We note that, although TAA growth was simulated under systolic loading, we selected maps of dilatation at diastolic pressure for training data, with the aim to emulate what can easily be computed from a standard clinical image (tending to capture anatomy around end-diastole). We generated 100 unique spatial distributions (Fig B in \hyperref[sec:SI]{S1 Supporting Information}), assigning 5 combinations of compromised elastic fiber integrity and mechanosensing to each profile, yielding 500 final dilatation-distensibility map pairs.

\subsubsection*{FE simulation \& dilatation/distensibility maps}

\subsection*{Numerical experiments}
Knowledge of the underlying mechanical and mechanobiological insults for a given TAA could provide valuable insight into aortic vulnerability and optimal interventions. With this goal of predicting the relative contributions of both initiating insults based only on imaging-derived quantities, we compared several candidate neural operator architectures (two forms of a Deep Operator Network, UNet, and Laplace Neural Operator, described below) and input data formats to establish a standard for handling subject-specific dilatation and distensibility information.

\subsubsection*{Input data formats}

\paragraph*{Dilatation only vs. dilatation \& distensibility}
In order to minimize radiation dose or imaging time, information is typically not acquired at multiple phases over the cardiac cycle; thus, only one image per clinical time point is available. Analogous to the current standard for surgical evaluation, we compared the performance of each network in predicting elastic fiber integrity and mechanosensing insults when trained only on dilatation maps (clinical standard) versus both dilatation and distensibility maps (requiring data over a cardiac cycle).

\paragraph*{Heat maps vs. grayscale maps}
We recently developed competing data preprocessing methods for constructing dilatation and distensibility maps. We interpolated $d$ and $\mathcal{D}$, evaluated on a nodal basis from the FE simulations in the $\theta$--$z$ domain, to $41 \times 41$ uniform grids either as the physical quantities (here named ``heat maps'') or after normalizing and converting to 8-bit integer ([0 255]) intensity fields (here named ``grayscale maps''), with dilatation maps sharing a preset range across all cases and distensibility maps each sharing a separate range across all cases. This approach was taken to allow straightforward integration with diverse neural operator architectures (discussed below), where heat maps are well-suited for feed-forward neural networks and grayscale maps for convolutional networks, as done previously \cite{Goswami2022}. Importantly, we also sought to evaluate our former best-performing architecture, based on grayscale map inputs, against alternative network designs.

\subsubsection*{Network architectures}

\begin{figure}[!t]
\centering
\includegraphics[width=0.95\textwidth]{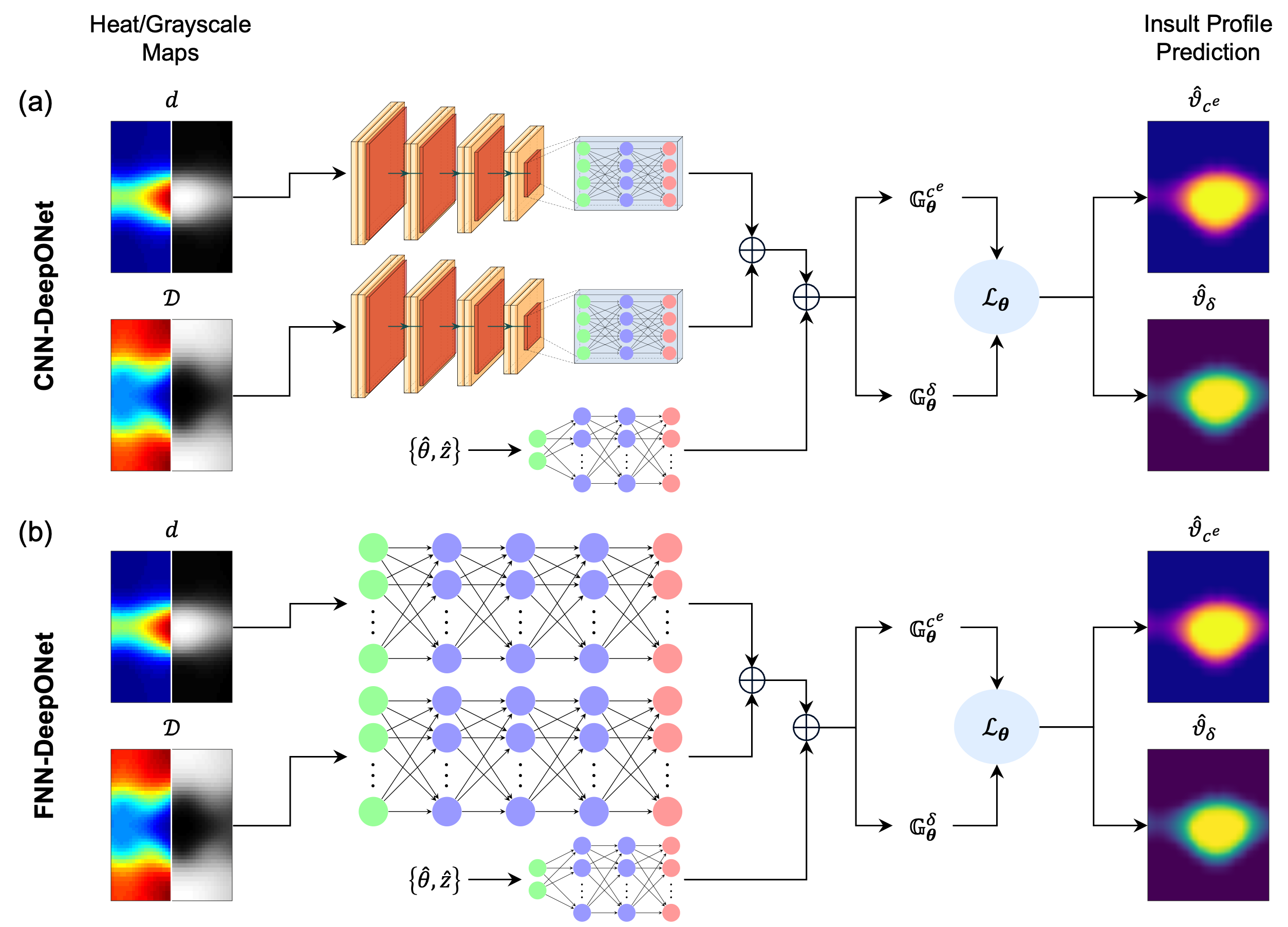}
\caption{{\bf Schematic representation of two DeepONet architectures.} Each branch net is a (a) CNN or (b) FNN that embeds the dilatation ($d$) and distensibility ($\mathcal{D}$) maps as either grayscale or heat maps. Each trunk net takes the coordinates $\{\hat\theta, \hat{z}\}$ to define the output domains of the corresponding insult contributor. The solution operators for each insult ($\mathbb G_{\bm{\theta}}^i$ ($i = c^e, \delta$)) are formed from element-wise dot products of the outputs of the branch and trunk networks, with shared learnable parameters ($\bm \theta$). Minimization of the loss function ($\mathcal L_{\bm{\theta}}$), defined as the combination of both operator outputs, determines the optimal parameters that enable estimation of the insult profiles and contributors ($\hat{\vartheta_i}$).}
\label{fig:DeepONet}
\end{figure}

\paragraph*{Deep Operator Network (DeepONet)}
DeepONet consists of branch networks to encode the input data and a trunk network to define the output domain, allowing a resolution-independent representation of input and output functions \cite{Lu2021}. Building on our previous framework, we compared the performance of two DeepONets that differ in branch net architecture, with $41 \times 41$ dilatation and distensibility maps serving as inputs, encoded via either fully connected convolutional neural networks (CNNs) (\autoref{fig:DeepONet}a) or feed-forward neural networks (FNNs) (\autoref{fig:DeepONet}b). In both DeepONets, the trunk net received initial positions of the aortic domain $\{\hat\theta, \hat{z}\}$ as input (as cylindrical coordinates), using an FNN. We defined separate solution operators for each insult contributor that served as inputs to the combined loss function with a single set of trainable parameters (mean squared error), minimization of which allowed estimation of the insult profiles for elastic fiber integrity and mechanosensing.

\begin{figure}[!t]
\centering
\includegraphics[width=0.95\textwidth]{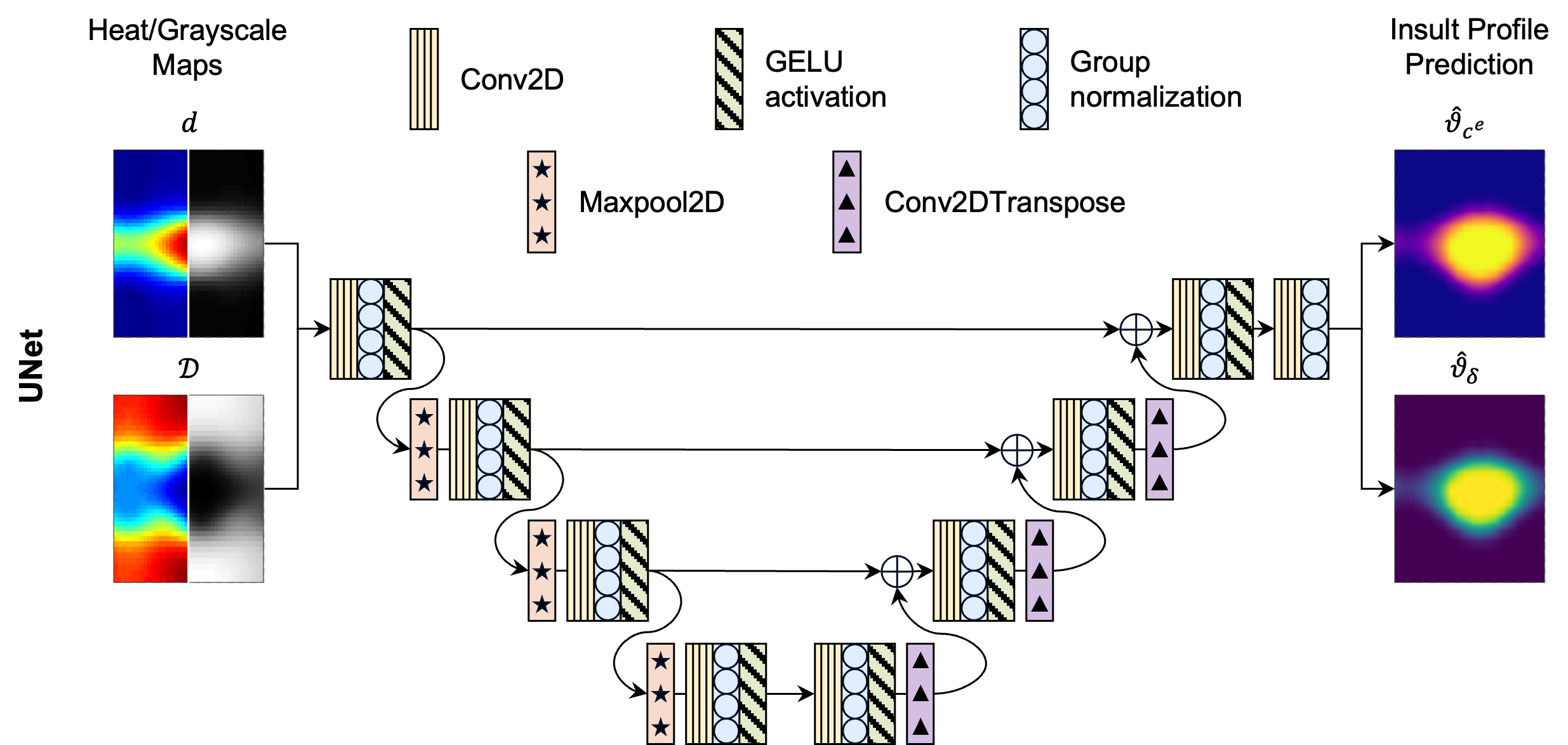}
\caption{{\bf Schematic representation of the UNet architecture.} Maps of dilatation ($d$) and distensibility ($\mathcal{D}$) are encoded via successive layers of two-dimensional convolution (Conv2D), group normalization, and Gaussian Error Linear Unit (GELU) activation. Down- and up-sampling the input by factors of 2 is achieved through two-dimensional max-pooling operations (Maxpool2D) and two-dimensional transpose convolutional operations (Conv2DTranspose), respectively. Finally, skip connections are implemented to propagate information from earlier layers.}
\label{fig:UNet}
\end{figure}

\paragraph*{UNet} 
UNets are U-shaped fully convolutional neural networks, originally developed for biomedical image segmentation applications \cite{ronneberger2015u}. UNets are also the building blocks of the diffusion models \cite{ho2020denoising} used in generative AI packages like DALL.E \cite{ramesh2021zero}. The extensive applications of UNets are rooted in their ability to learn the mapping from input to output signals through latent representations at varying degrees of coarseness, with coarser representations responsible for learning low-frequency components of the solution and finer representations responsible for high-frequency components. The multigrid representation learning makes the UNet effective in extracting spatiotemporal correlations entangled in the solutions of PDEs \cite{gupta2022towards, ovadia2023ditto}, subsequently leading to widespread applications from materials science \cite{oommen2024rethinking} to turbulence modeling \cite{oommen2024integrating}. We provided the dilatation and distensibility information as $41 \times 41$ inputs to a UNet that learned the mapping to the corresponding insult profiles (\autoref{fig:UNet}).

\begin{figure}[!t]
\centering
\includegraphics[width=0.95\textwidth]{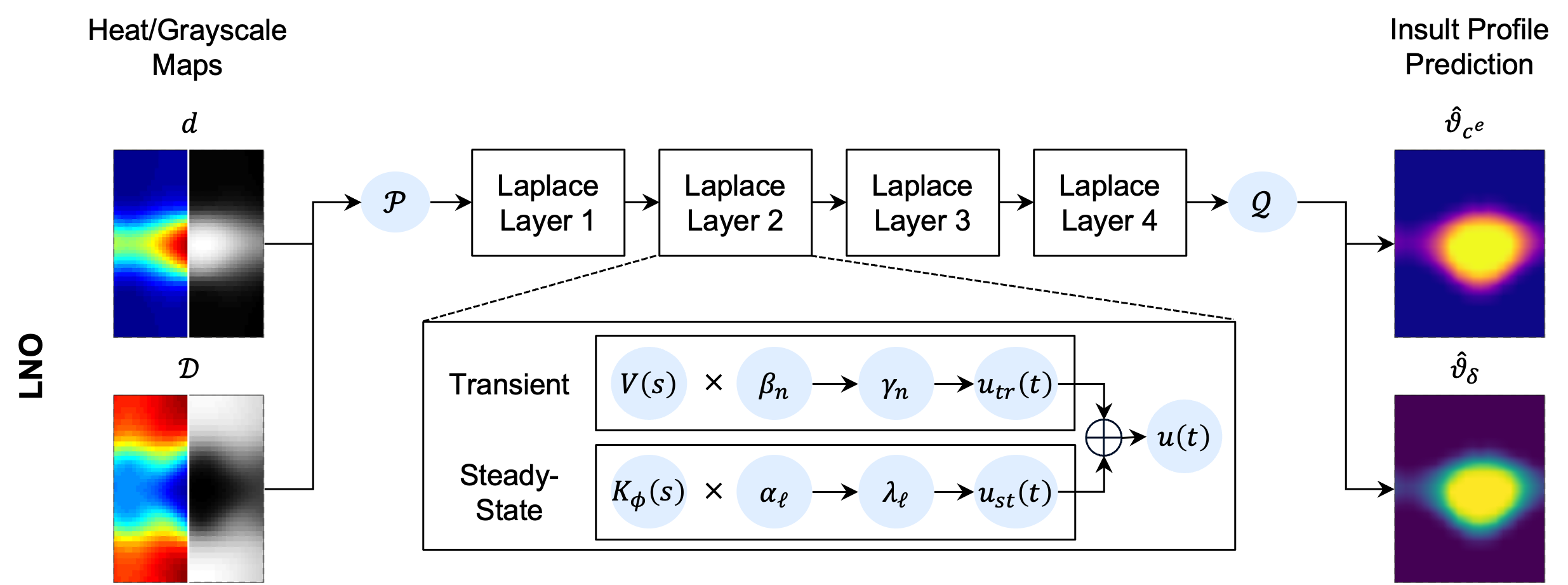}
\caption{{\bf Schematic diagram of the LNO architecture.} The dilatation ($d$) and distensibility ($\mathcal{D}$) are lifted to a higher dimension via shallow neural network $\mathcal{P}$, to which Laplace layers are applied, each yielding the output $u(t) =f[ K_\phi(s)V(s)]$, where $K_\phi(s)$ is the Laplace transform of the kernel integral transformation and $V(s)$ is the Laplace transform of the lifted input function. Each layer contains pole-residue methods to obtain the transient and steady-state responses ($u_{tr}(t)$ and $u_{st}(t)$, respectively), with residues $\beta_n$, coefficients $\alpha_\ell$, and response poles $\gamma_n$ and $\lambda_\ell$. Finally, the outputs (insult profiles) are projected back into the target dimension using the shallow network $\mathcal{Q}$.}
\label{fig:LNO}
\end{figure}

\paragraph*{Laplace Neural Operator (LNO)}
The Laplace Neural Operator is an architecture that performs operator learning in the Laplace domain for solving ordinary and partial differential equations \cite{cao2024laplace}. This architecture leverages the solution of PDEs represented by the integral of Green's function and the kernel integral is transformed and calculated in the Laplace domain. A key innovation in LNO is its Laplace layer, which employs an analytical pole-residue operation to establish a physically interpretable and meaningful mapping between the input and output functions in the Laplace domain. By independently learning the steady-state response, transient response with zero initial conditions, and transient response under nonzero initial conditions, LNO effectively captures true system dynamics. This enables it to achieve better approximation accuracy compared to other neural operators for extrapolation circumstances and dynamical systems. In this study, four Laplace layers were chosen, with width and modes for each layer being 32 and 8, respectively (\autoref{fig:LNO}).

\subsection*{Training}
The 500 FE simulations were randomly categorized into 450 training and 50 testing samples. We then evaluated the prediction accuracy of each architecture-input data combination with a relative $\mathcal{L}_2$ error computed over all testing samples, with separate errors for compromised elastic fiber integrity and mechanosensing. We also computed point-wise absolute errors in each predicted insult profile ($\hat{\vartheta}_i - \vartheta_i$, $i = c^e, \delta$, where $\hat{\vartheta}_i$ is the predicted insult profile and $\vartheta_i$ is the ground truth) to assess the ability of each network to reproduce the insult spatial distributions.

For the DeepONets, the initial architecture, learning rate, and number of training epochs were selected based on commonly used configurations reported in prior work \cite{Goswami2022}. Building on this baseline, we conducted a grid search over activation functions, network depth, and width, while monitoring loss curves to ensure convergence and mitigate overfitting. The final architecture was chosen based on the configuration that yielded the lowest validation loss within a predefined number of epochs (200,000), resulting in an Adam optimizer with the learning rate set at 0.001. In the cases of dilatation-only training, the branch net encoding distensibility was omitted from both architectures. For the UNet, the batch size was set to 250 and trained for 100,000 epochs. Following \cite{gupta2022towards}, we used GELU activations \cite{hendrycks2016gaussian} and a cosine-annealing-based learning rate scheduler with an initial learning rate of 0.0001. For the LNO, the dilatation and distensibility information were provided as the inputs with dimensions $\{20, 41, 41, 2\}$, and the two insults were the outputs with dimensions $\{20, 41, 41, 2\}$, where 20 is the batch size, 41 is the 2D resolution of the map, and 2 is the number of channels. Both the inputs and outputs were normalized by min-max normalization during training. The relative $\mathcal{L}_2$ error between the predicted insults and the true insults was used as the loss function. We first adopted commonly used settings from similar works \cite{cao2024laplace} and performed a series of experiments to ensure convergence without overfitting, yielding 10,000 epochs and an exponential decay learning rate schedule with an initial learning rate of 0.001. The best model during these 10,000 epochs was chosen as the final surrogate model. Relevant network parameters are summarized in \autoref{tab:training}.

\begin{table}[!t]
\centering
\footnotesize
\begin{threeparttable}
\caption{Parameters for all network architectures. Note that the number of weight updates does not correspond to the number of epochs.}
\label{tab:training}
{
\begin{tabular}{l c c c c c}
\hline
Network & \makecell{\# Trainable\\Parameters} & FLOPS$^\ast$ & \makecell{Activation\\Function} & \makecell{Learning\\Rate} & \makecell{\# Weight\\Updates} \\
\hline
CNN-DeepONet & $1.95\times10^{6}$ & 24.2e+7 & SiLU   & 1e--3$^\dagger$ & 200,000\\
FNN-DeepONet & $0.31\times10^{6}$ & 23.6e+7 & SiLU   & 1e--3$^\ddagger$ & 200,000\\
UNet         & $2.04\times10^6$   & 5.27e+7 &  GELU  & 1e--4$^\ddagger$ & 200,000\\
LNO          & $0.34\times10^6$   & 1.44e+7 & $\sin$ & 1e--3$^\dagger$ & 225,000\\
\hline
\end{tabular}
\begin{tablenotes}
    \item[$^\ast$] Number of floating point operations per function evaluation. \vspace{-2pt}
    \item[$^\dagger$] Constant learning rate. \vspace{-2pt}
    \item[$^\ddagger$] Cosine annealing learning rate scheduler.
\end{tablenotes}
}
\end{threeparttable}
\end{table}

\section*{Results}
\subsection*{Effects of insult contributors on dilatation and distensibility}
FE simulations of TAAs initiated from combinations of compromised elastic fiber integrity and mechanosensing equilibrated with normalized inner radius ranging from 0.973 (at the vessel ends) to 1.657 (at the apex of the insult region), with a mean maximum dilatation of 1.496 $\pm$ 0.0476. Insult profiles having a greater span in the circumferential direction generated greater levels of dilatation, consistent with previous investigations \cite{Goswami2022}. The initial non-aneurysmal aorta exhibited a distensibility of 0.05442 (i.e., 5.4\%), while aneurysmal dilatation decreased distensibility within the insult region, with a mean minimum distensibility of 0.0344 $\pm$ 0.0038 (i.e., 3.4\%), corresponding to a $\sim$37\% reduction associated with the $\sim$50\% dilatation.

We further observed differential effects on the maximum dilatation and minimum distensibility depending on the dominating contributor to the overall insult. Although all combinations of compromised elastic fiber integrity and mechanosensing resulted in significant dilatation and decreased distensibility, mechanosensing-dominated insults tended to have consistently high dilatation, while elastic fiber integrity-dominated insults corresponded to lower distensibility within the insult region. Regions opposite the location of minimum distensibility also exhibited reduced distensibility despite not directly experiencing a prescribed insult, which occurred for all combined insults considered. Representative dilatation and distensibility profiles corresponding to the full range of combined insult magnitudes can be seen in Fig A in \hyperref[sec:SI]{S1 Supporting Information}.

\subsection*{Overall prediction accuracy}

\begin{figure}[!t]
\centering
\includegraphics[width=\textwidth]{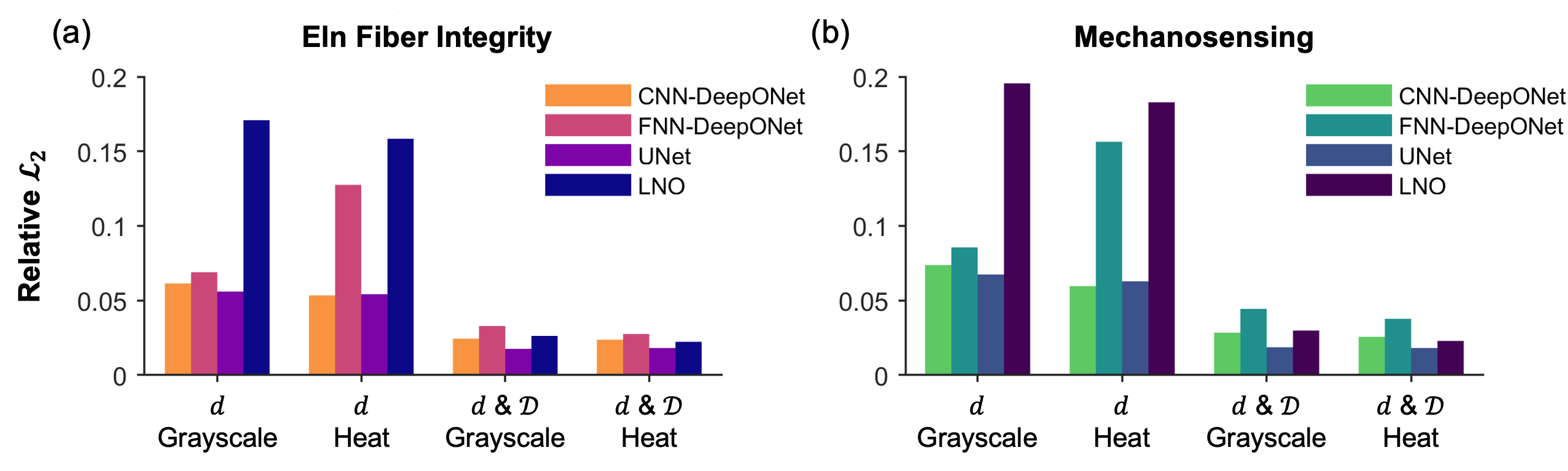}
\caption{{\bf Performance of four different neural networks in predicting insult contributors to aneurysmal dilatation.} Relative $\mathcal{L}_2$ errors are reported separately for compromised (a) elastic (eln) fiber integrity and (b) mechanosensing over all testing cases considered: dilatation ($d$) only in grayscale and heat map formats, and dilatation and distensibility ($d$ \& $\mathcal{D}$) in grayscale and heat map formats.}
\label{fig:all-errors}
\end{figure}

\begin{table}[!t]
\centering
\caption{Relative $\mathcal{L}_2$ errors for all network-input data combinations predicting combined insult contributors shown in \autoref{fig:all-errors}, evaluated over all testing cases. The best results (lowest $\mathcal{L}_2$ error) are highlighted by boldface in both cases.}
\resizebox{\textwidth}{!}{
\def\arraystretch{1.1}
\begin{tabular}{ l >{\centering\arraybackslash}p{3cm} >{\centering\arraybackslash}p{3cm} >{\centering\arraybackslash}p{3cm} >{\centering\arraybackslash}p{3cm} }
\hline
& \multicolumn{4}{c}{Overall Relative $\mathcal{L}_2$ Error} \\
& $d$ Grayscale & $d$ Heat & $d$ \& $\mathcal{D}$ Grayscale & $d$ \& $\mathcal{D}$ Heat \\
\hline
\bf{Eln Fiber Integrity} & & & & \\
\quad CNN-DeepONet & 0.0614 & 0.0534 & 0.0244 & 0.0237 \\
\quad FNN-DeepONet & 0.0689 & 0.1275 & 0.0329 & 0.0276 \\
\quad UNet & 0.0560 & 0.0542 & \bf{0.0176} & 0.0182 \\
\quad LNO & 0.1708 & 0.1584 & 0.0263 & 0.0223 \\
\hline
\bf{Mechanosensing} & & & & \\
\quad CNN-DeepONet & 0.0738 & 0.0596 & 0.0284 & 0.0257 \\
\quad FNN-DeepONet & 0.0856 & 0.1564 & 0.0444 & 0.0377 \\
\quad UNet & 0.0675 & 0.0628 & 0.0187 & \bf{0.0182} \\
\quad LNO & 0.1955 & 0.1829 & 0.0299 & 0.0229 \\
\hline
\end{tabular}}
\label{tab:all-rel-l2}
\end{table}

Following training with the same 450 simulations, the four neural operator architectures were used to predict the remaining 50 insult fields of compromised elastic fiber integrity and dysfunctional mechanosensing based on inputs of dilatation only or dilatation and distensibility, each provided as either grayscale or heat maps. Most networks were able to predict the initiating insults within 10\% relative $\mathcal{L}_2$ error for both elastic fiber integrity (\autoref{fig:all-errors}a) and mechanosensing (\autoref{fig:all-errors}b). In particular, CNN-DeepONet and UNet architectures consistently predicted both insult profiles with approximately 5\% error, even when trained on dilatation only. However, the relative $\mathcal{L}_2$ error for the LNO in dilatation only cases exceeded 15\%, and the FNN-DeepONet in the dilatation only heat map case rose above 10\% for both insult contributors. \autoref{tab:all-rel-l2} shows the corresponding prediction errors.

\begin{figure}[!t]
\centering
\includegraphics[width=\textwidth]{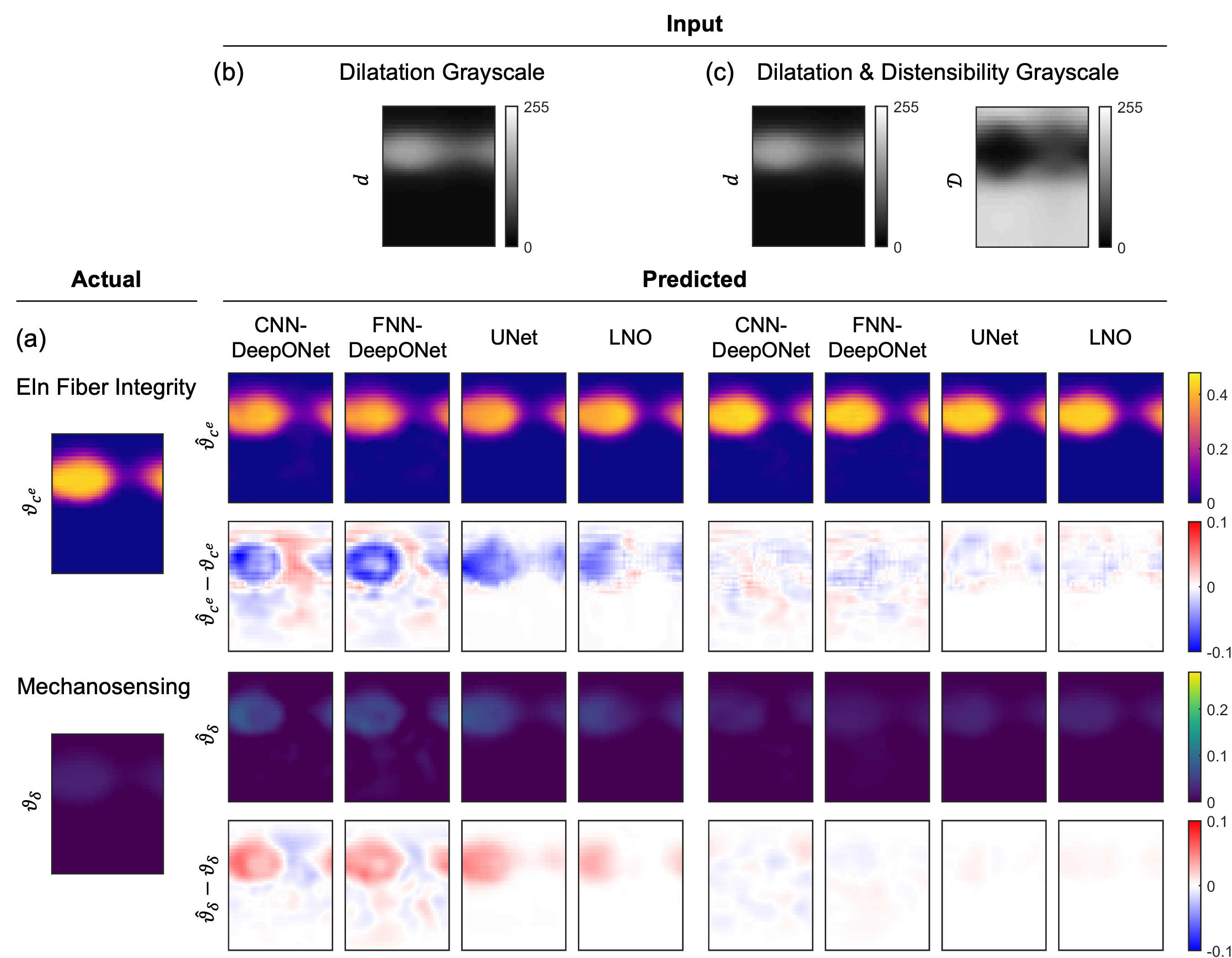}
\caption{{\bf Predictions from all four architectures for an elastic fiber integrity-dominated combined insult.} (a) The ground truth combined insult field consisted of contributions of both compromised elastic fiber integrity ($\vartheta_{c^e}$) and dysfunctional mechanosensing ($\vartheta_{\delta}$) superimposed in the FE simulation to generate the TAA with dilatation and distensibility profiles shown in (b--c). Predictions ($\hat{\vartheta}_i$) and absolute errors ($\hat{\vartheta}_i - \vartheta_i$ ($i=c^e,\delta$)) are shown for CNN-DeepONet, FNN-DeepONet, UNet, and LNO trained on (b) dilatation grayscale maps only and (c) dilatation and distensibility grayscale maps.}
\label{fig:ef-allnets-gray}
\end{figure}

\begin{figure}[!t]
\centering
\includegraphics[width=\textwidth]{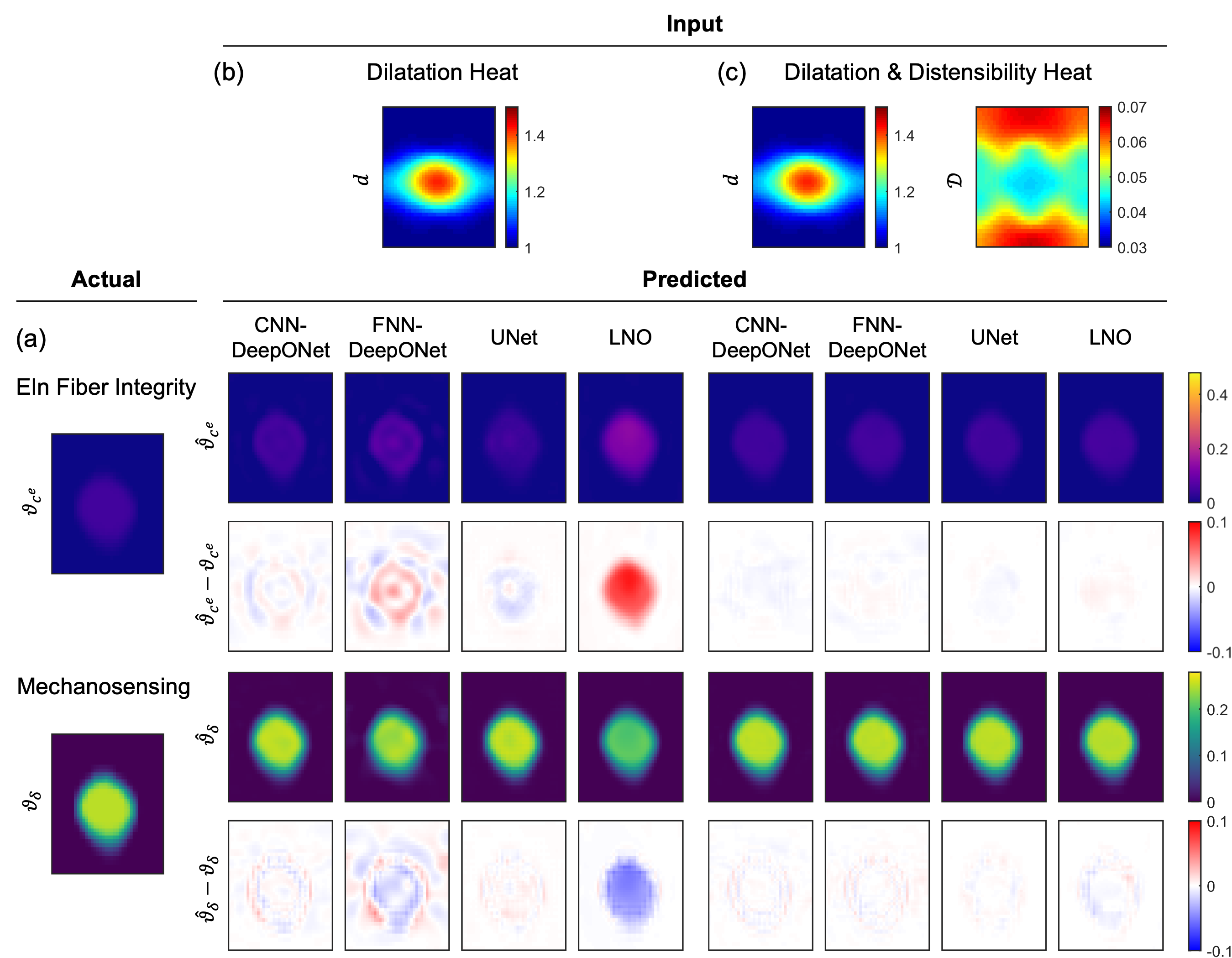}
\caption{{\bf Predictions from all four architectures for a mechanosensing-dominated combined insult.} Similar to \autoref{fig:ef-allnets-gray} except for heat map inputs. (a) The ground truth combined insult field consisted of contributions of both compromised elastic fiber integrity ($\vartheta_{c^e}$) and dysfunctional mechanosensing ($\vartheta_{\delta}$) superimposed in the FE simulation to generate the TAA with dilatation and distensibility profiles shown in (b--c). Predictions ($\hat{\vartheta}_i$) and absolute errors ($\hat{\vartheta}_i - \vartheta_i$ ($i=c^e,\delta$)) are shown for CNN-DeepONet, FNN-DeepONet, UNet, and LNO trained on (b) dilatation heat maps only and (c) dilatation and distensibility heat maps.}
\label{fig:ms-allnets-heat}
\end{figure}

Figs \ref{fig:ef-allnets-gray} and \ref{fig:ms-allnets-heat} show representative examples of insult profile predictions for each network. Between the DeepONet designs, the FNN-DeepONet exhibited worse overall performance than the CNN-DeepONet, with the greatest difference in the dilatation-only heat map case. The CNN-DeepONet was able to achieve a similar prediction accuracy as the UNet when both were trained with dilatation only, whereas its performance appeared similarly competitive with the LNO when trained with both dilatation and distensibility. LNO, while better than both DeepONets at predicting the insult profiles when trained on both dilatation and distensibility, showed the worst prediction error in dilatation-only cases. Overall, UNet achieved the greatest prediction accuracy, particularly with dilatation and distensibility input data (Figs C and D in \hyperref[sec:SI]{S1 Supporting Information} show additional comparisons).

The greatest prediction error concentrated in the insult region with dilatation-only training; however, both DeepONet frameworks further yielded noticeable prediction errors away from the insult region, while UNet and LNO frameworks represented the non-insult region with high accuracy. Testing samples in which contributions of elastic fiber integrity and mechanosensing insults were more evenly balanced tended to be predicted with greater accuracy, while combined insults dominated by one factor were predicted less well.

\subsection*{Effects of distensibility maps}
The inclusion of distensibility data in training reduced the prediction errors for all four networks by roughly a factor of 2 (\autoref{fig:all-errors}). As can be seen in \autoref{fig:ef-allnets-gray} and \autoref{fig:ms-allnets-heat}, the prediction errors for both insult contributors effectively vanished upon the use of distensibility maps, especially for the high prediction errors of the FNN-DeepONet and LNO, and the network performances became comparable to one another. Nevertheless, there remained persistent small prediction errors with similar patterning in the dilatation-only cases, with DeepONet errors throughout the model domain and UNet and LNO errors clustered around the insult region.

\subsection*{Effects of input data format}
\begin{figure}[!ht]
\centering
\includegraphics[width=\textwidth]{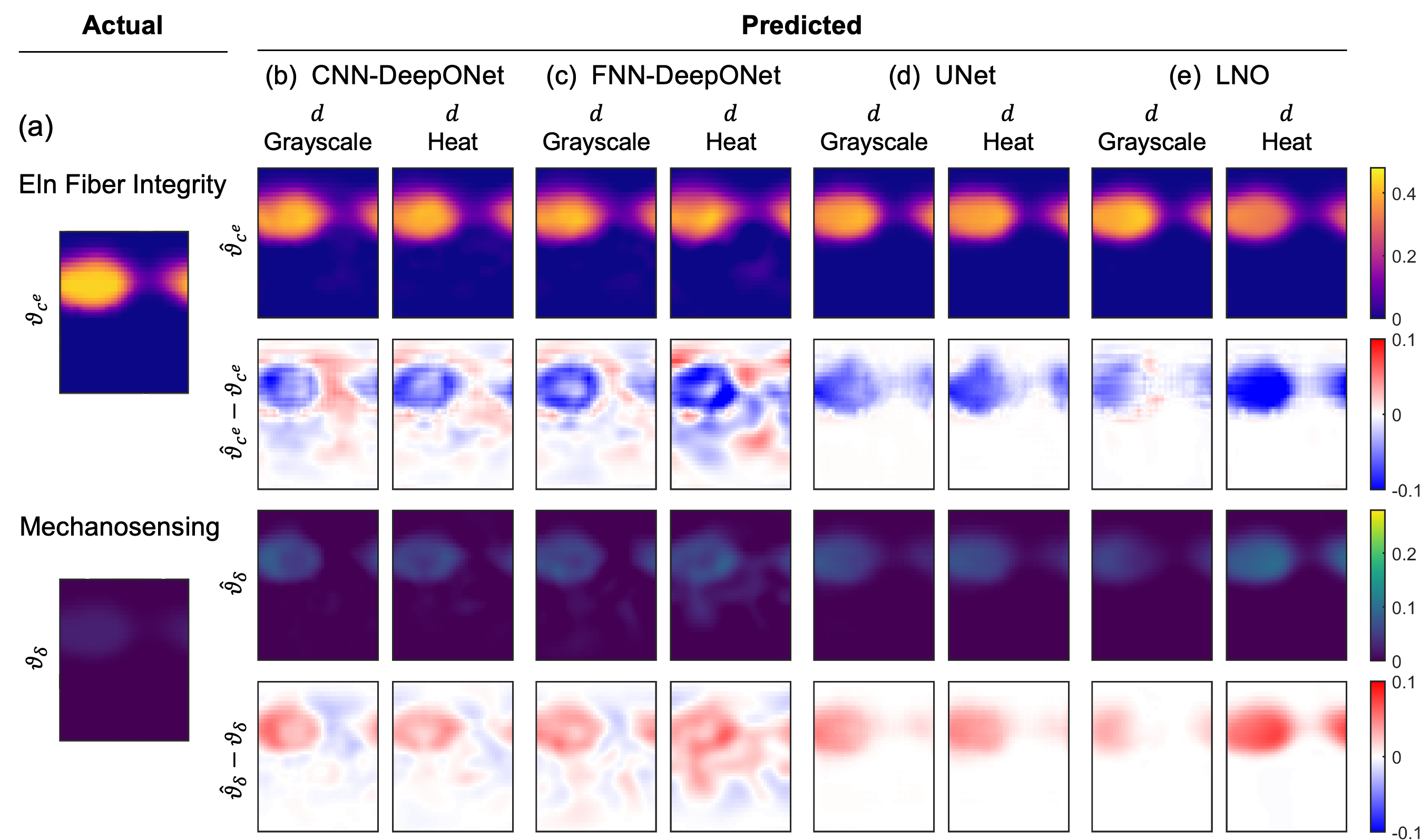}
\caption{{\bf Effects of grayscale versus heat map data inputs.} All networks were trained on dilatation ($d$) maps only. (a) Ground truths for elastic fiber integrity and mechanosensing contributions. Predictions ($\hat{\vartheta}_i$) and computation of absolute errors ($\hat{\vartheta}_i - \vartheta_i$ ($i=c^e,\delta$)) are shown for (b) CNN-DeepONet, (c) FNN-DeepONet, (d) UNet, and (e) LNO.}
\label{fig:all-cases-dil}
\end{figure}

Interestingly, use of grayscale versus heat maps for training did not significantly impact the predictions of the networks based on CNNs (\autoref{fig:all-cases-dil}a--b) or the LNO; for most networks, we observed only modest gains in prediction accuracy when using heat maps, and UNet performance was virtually the same regardless of data format (\autoref{fig:all-cases-dil}e--f). Yet, heat maps slightly improved the performance of the FNN-DeepONet in dilatation-distensibility cases (\autoref{fig:all-errors}), but their use in dilatation-only cases resulted in worse prediction errors (\autoref{fig:all-cases-dil}c--d).

\section*{Discussion}
In previous work, we established the use of DeepONets comprised of CNN-based branch networks to estimate insults to the aortic wall resulting from either compromised elastic fiber integrity or dysfunctional mechanosensing based on dilatation and distensibility information. We found that, for single-contributor insults with arbitrary spatial distributions, full-field measurements of dilatation and distensibility in grayscale map format led to more robust predictions compared to sparse sensor point grids, demonstrating the gain in prediction accuracy with two-dimensional knowledge of the geometry and mechanical properties of the aneurysmal aorta rather than measuring only at sparse locations around the point of maximum dilatation \cite{Goswami2022}. Herein, we extended this prior approach by generating new synthetic data that superimposes multiple insults to both mechanical properties and mechanobiological processes, while contrasting multiple neural network architectures and formats for data input. Combined with parameterization of the initial model to a diseased murine ascending aorta, this allowed the synthetic data to further close the gap between hypothetical \emph{in silico} and \emph{in vivo} cases. Moreover, to bracket the potential loss in predictive accuracy when the gold standard of input data cannot be reached, we investigated an intermediate case in which only dilatation measurements were available to train the model. Finally, with recent advancements in neural network applications to biomedical image analysis, we assessed the performance of UNets and LNOs as alternative approaches in performing the same tasks.

\subsection*{UNet as the preferred architecture for estimating combined insults}
Reasonable predictions of combined insults were achieved with all networks considered in this investigation. Our previous DeepONet-based architecture continued to perform well for multi-contributor TAAs, including an additional framework comprised of FNNs, confirming our initial choice of network. Nevertheless, these architectures were still prone to modest prediction errors throughout the model domain, even far from the actual insult region (this can be seen in Fig E in \hyperref[sec:SI]{S1 Supporting Information}, where despite the good predictions achieved by all networks, spatial distributions of error varied widely by network type). Alternatively, both UNet and LNO predictions consistently captured the boundaries of the insult region with much greater accuracy than the DeepONets, owing to their ability to capture low-order and high-order features in the input data (in contrast, fully connected CNNs are considered better with texture mapping than with feature detection in encoding images). Even significant UNet and LNO errors remained concentrated at the insult region. This difference is especially prominent in dilatation-only testing samples in which all networks yielded poor predictions (\autoref{fig:ef-allnets-gray}). LNO performed the worst in these scenarios, though this is due to incorrectly estimating the magnitude of insult contributors and not because of failing to correctly identify the spatial distribution of the insult. We emphasize that, for assessing the functional state of the aorta, correct prediction of the insult magnitudes is paramount, especially if the clinical focus is the region of greatest dilatation.

Because predictions away from the aneurysm apex may be unrealistic to validate in clinical cases, we also reevaluated the error estimates on only the regions where the normalized insult was 50\% or above ($\vartheta^\ast \geq 0.50$), concentrating on the most affected regions in each testing sample. Even after this filtering, the overall performances of each network exhibited the same trends as when evaluated on the whole domain, with one exception that UNets trained on heat maps rather than grayscale maps became the best performing architecture (Table B in \hyperref[sec:SI]{S1 Supporting Information}). Overall, these results suggest for dilatation-only scenarios, both CNN-DeepONet and UNet could be appropriate choices, with the caveats of small, diffuse prediction errors arising from the DeepONet and the requirement that input data be interpolated onto square domains. In dilatation-distensibility cases, UNet and LNO outperformed the DeepONet-based frameworks in predicting the combined insult profiles, with UNet exhibiting the best prediction accuracy overall regardless of the input data format. Therefore, for predictions based on two-dimensional full-field maps, UNet is the preferred choice due to its accuracy and multi-scale processing capabilities, even if distensibility is not available.

\subsection*{Importance of distensibility in mixed insult data}
We observed correlations between maximum dilatation and minimum distensibility depending on the relative contributions of elastic fiber integrity and mechanosensing insults. All combined insults decreased distensibility, though arising from different causes. In the case of compromised elastic fiber integrity, the prescribed reduction in mechanical properties of the elastin-dominated extracellular matrix resulted in both a transferal of load from the more compliant elastin to the stiffer collagen as well as further simulation of production of collagen fibers; this double-hit contributed to the decreased distensibility. On the other hand, dysfunctional mechanosensing left the properties of the elastic fibers intact while governing the turnover of smooth muscle cells and especially collagen fibers to stiffen the wall. Being that elastic fibers play a dominant role in the ability of the aorta to distend with blood pressure changes, it is not surprising that the most dramatic effects are observed in elastic fiber integrity-dominated insults. This can be seen in Fig F in \hyperref[sec:SI]{S1 Supporting Information}, in which synthetic TAAs that exhibit mostly the same average dilatation distributions fall into distinct distensibility patterns as a function of elastic fiber loss. We hypothesize that this differentiation in distensibility profile is what allows the networks to achieve accurate predictions when provided distensibility maps in either grayscale or heat map formats, as well as why dilatation-only cases proved more difficult to predict well.

Finally, examining the computed stress distributions within the synthetic data reveals dramatically different profiles depending on the dominating contributor (Fig G in \hyperref[sec:SI]{S1 Supporting Information}). In particular, mechanosensing-dominated combined insults associated with significantly higher circumferential ($\sigma_{\theta\theta}$) and axial ($\sigma_{zz}$) stresses within the aneurysmal region. Estimating the intramural shear stress as $(\sigma_{\theta\theta} - \sigma_{zz})/2$, which could serve as an analog for understanding vulnerability to dissection, is shown to be sensitive to mechanosensing dysfunction. This underscores that the mechanisms driving aortic dilatation may have a profound impact on the mechanical stability of the aneurysm while presenting nearly identically based on geometry alone. We thus reiterate that the ability to distinguish underlying contributors to TAA may provide critical insight into the potential mechanical strength of the aortic wall within the aneurysmal region.

\subsection*{Limitations}
This study focused on combined insults of compromised elastic fiber integrity and dysfunctional mechanosensing in a model of a single initial geometry and material parameter set. Other studies suggest that additional effects could be considered, including altered mechanoregulation of matrix and compromised collagen cross-linking, in the synthetic data. We submit nonetheless that this work constitutes a viable potential pipeline demonstrating that dilatation and distensibility maps are essential to accurately estimate superimposed insults. Additionally, future work will need to incorporate increased variability in vessel dimensions, material properties, and loading conditions from diverse experimental investigations.

Next, the FE simulation platform used a computationally efficient implementation of equilibrated aortic growth and remodeling, yielding TAA geometries considered to be mechanobiologically stable, meaning no further growth of the TAA would be expected. Although this may be appropriate for a portion of TAA patients, these synthetic datasets do not address a TAA that would undergo unstable growth or develop a dissection or rupture. Furthermore, uniform cylindrical initial geometries were used to generate dilatation and distensibility maps; while effort was made to simulate TAAs of appropriate size for the initial vessel, the curvature of the ascending aorta was not considered. This simplification in original geometry allowed us to assess effects of combined insults, the utility of possible down-sampling, and so on, while targeting a level of dilatation (1.5-fold increase) that is clinically important independent of effects of curvature of the ascending aorta. Therefore, future work will need to consider both stable and unstable growth of TAAs based on combined insults applied to image-based aortic geometries, in which the UNet and LNO will be expected to have particular relevance. Looking to the future, we note that, if provided synthetic data of a longitudinal, time-evolving nature, different performances for these networks may reveal an alternative preferred choice of network, given that LNO is better suited for temporal applications and DeepONet is not as well suited.

Finally, there remains a compelling need to adapt this synthetic data pipeline to incorporate experimental data from actual human TAAs. With recent advancements in rapid three-dimensional reconstruction of the aorta from medical images \cite{Burris2022}, obtaining biomechanical properties from human samples \cite{Sommer2018}, and using machine learning surrogate models to augment or complement high-cost biomechanical simulations \cite{Motiwale2024}, the potential to generate robust training data from clinically relevant growth and remodeling models may soon be a reality, at which point these pre-trained neural operators can help improve clinical decision-making.

\section*{Conclusion}
In summary, using machine learning approaches for accurate prediction of mechanobiological insults that give rise to similarly sized TAAs promises to better discriminate mechanical vulnerability, thereby helping to inform subject-specific clinical treatment planning. Toward this end, we performed a systematic comparison of multiple neural networks in predicting the spatial distribution of synthetic multi-contributor TAA initiators when trained on dilatation information alone versus both dilatation and distensibility. Our findings highlight that:

\begin{itemize}
    \item Full-field measurement of the ascending aortic domain facilitates localized assessment of the underlying aortic insult, advancing beyond the current clinical standard of evaluating a singular maximum aortic size.
    \item Localized dilatation fields have substantial predictive value in estimating mechanisms of TAA progression; however, corresponding knowledge of local distensibility is essential to accurately determine the relative contributions of multiple disease drivers.
    \item UNet-based neural networks are an efficient and promising tool for performing these predictions based on image-derived aortic quantities.
\end{itemize}

Ultimately, as progress continues in machine learning-assisted automatic segmentation of medical images and generation of biomechanical FE models, the prospect of obtaining these scalar-field measurements in a rapid, patient-specific manner could yield a minimally invasive, image-based approach for improved TAA risk assessment and management.

\section*{Data accessibility}
The data and scripts used in this study are available through GitHub (\url{https://github.com/Centrum-IntelliPhysics/Mixed-Insults-Aortic-Aneurysm}).

\section*{Ethics statement}
The data used to parameterize the simulations in this study was obtained from previously published work \citep{Cavinato2021} that was reviewed and approved by the Institutional Animal Care and Use Committee of Yale University.

\section*{Acknowledgments}
This work was funded in part by grants from the National Institutes of Health (R01 HL168473 to RA and GK, R01 HL163811 to Jonathan Weinsaft). The authors would like to acknowledge the computing support provided by the Advanced Research Computing at Hopkins (ARCH) core facility at Johns Hopkins University and the Rockfish cluster, as well as the computational resources and services at the Center for Computation and Visualization (CCV) at Brown University, where all experiments were carried out. The ARCH core facility (\url{rockfish.jhu.edu}) is supported by the National Science Foundation (OAC1920103).








\bibliographystyle{abbrvnat}
\bibliography{refs}


\newpage
\beginsupplement
\setcounter{section}{0}
\setcounter{subsection}{0}
\section*{Supplemental material}
\label{sec:SI}

\section*{Constitutive relations}
We employed a well-established framework for aortic growth and remodeling (G\&R) based on the notion of mechanobiological homeostasis in a healthy aorta in maturity \cite{Murtada2021}, wherein aortic cells modulate the production and removal of extracellular matrix components in response to perturbations in order to maintain intramural stress near homeostatic ``set-points'' $\sigma_o$ \cite{Humphrey2021} The aortic wall was modeled as a constrained mixture of constituents $\alpha$ defining its material properties, namely elastin-dominated matrix ($\alpha = e$), smooth muscle cells ($\alpha = m$), and collagen-dominated matrix ($\alpha = c$). Importantly, the latter two constituents are capable of turnover via growth (changes in mass) and remodeling (changes in structure) in response to deviations in stress from normal $\Delta \sigma = ((1-\delta) \sigma - \sigma_o)/\sigma_o$, where $\delta$ captures mechanosensing capability and $\sigma = \rm{tr} \, \boldsymbol{\sigma}/3$ is a scalar representation of the current intramural (Cauchy) stress defined as

\begin{equation}
    \boldsymbol{\sigma} = 2 \mathbf{F}(\partial W/\partial \mathbf{C}) \mathbf{F}^\textrm{T}/\det \mathbf{F},
\end{equation}\smallskip

\noindent with $\mathbf{F}$ the deformation gradient tensor, $\mathbf{C}$ the right Cauchy-Green tensor, and $W$ the constitutive relation for total stored energy. $W$ may be defined conceptually as a sum of constituent-specific stored energies ($W = \sum_{\alpha = e,c,m} \phi^\alpha \hat{W}^\alpha$), which we defined separately for elastin and smooth muscle cells and collagen as

\begin{equation}
    \hat{W}^e = c^e (I_1^e - 3) \quad \hat{W}^{m,c} = \frac{c_1^{m,c}}{4c_2^{m,c}} \Big( \exp \big( c_2^{m,c}(I_4^{m,c} - 1)^2 \big) - 1 \Big)
\end{equation}\smallskip

\noindent with associated mass fractions $\phi^\alpha$. 

In this study, we assumed that the stimulus functions were not responsive to changes in wall shear stress owing to both endothelial dysfunction and smooth muscle phenotypic modulation associated with thoracic aortic aneurysm (TAA). Geometric, material, and turnover parameters used in this study are summarized in \autoref{tab:params}.

\begin{table}[!ht]
\centering
\caption{{\bf Baseline material and G\&R parameters.} Superscripts $e, m, c$ denote elastin-dominated matrix, smooth muscle, and collagen-dominated matrix, respectively; super/subscripts $r, \theta, z, d$ denote radial, circumferential, axial, and symmetric diagonal directions, respectively. Data are based on studies of mouse models of thoracic aortopathy \cite{Cavinato2021, Li2023}.}
\resizebox{\textwidth}{!}{
\def\arraystretch{1.1}
\begin{tabular}{ l >{\centering\arraybackslash}p{3cm} c }
\hline
& Parameter & Value \\
\hline
Inner radius, thickness, length & $r_o, h_o, l_o$ & 0.808 mm, 0.041 mm, 10 mm \\
Elastin, smooth muscle, collagen mass fractions & $\phi_o^e, \phi_o^m, \phi_o^c$ & 0.354, 0.298, 0.348 \\
Collagen orientation fractions & $\beta^\theta, \beta^z, \beta^d$ & 0.077, 0.035, 0.888 \\
Diagonal collagen orientation & $\alpha_{0o}$ & $50.3^\circ$ \\
Elastin material parameters & $c^e$ & 92.6 kPa \\
Smooth muscle material parameters & $c_1^m, c_2^m$ & 0.32 kPa, 31.1 \\
Collagen material parameters & $c_1^c, c_2^c$ & 4446.5 kPa, 2.45 \\
Elastin deposition stretches & $G_r^e, G_\theta^e, G_z^e$ &  $1/(G_\theta^eG_z^e)$, 2.11, 2.03 \\
Smooth muscle and collagen deposition stretches & $G^m, G^c$ & 1.20, 1.06 \\
Mechanosensing & $\delta$ & 0 \\
Smooth muscle-to-collagen turnover ratio & $\eta$ & 1.0 \\
Shear-to-intramural gain ratio & $K_{\tau_w}/K_\sigma$ & 0.2 \\
\hline
\end{tabular}}
\label{tab:params}
\end{table}

\section*{Insult profiles generated via Gaussian random fields}

As described previously \cite{Goswami2022}, we used Gaussian random fields (GRFs) to define randomly distributed insult profiles throughout the aortic ($z$--$\theta$) domain. Latent profiles $\vartheta^\ast$ were sampled according to

\begin{equation}
    \vartheta^\ast (z_o, \theta_o) \sim \mathcal{G} \big( \mu(z_o, \theta_o), \kappa(z_o, \theta_o, z_o', \theta_o') \big),
\end{equation}\smallskip

\noindent where $\mu(z_o, \theta_o)$ and $\kappa(z_o, \theta_o, z_o', \theta_o')$ are the mean and covariance between points $(z_o, \theta_o)$ and $(z_o', \theta_o')$, with $o$ denoting the reference configuration. The mean and covariance can be controlled by the user to specify the overall approximate insult surface fraction $\varphi$, boundary softness of the insult region $\epsilon$, and length scale along the circumferential and axial directions, $L_\theta$ and $L_z$, respectively. The mean and variance of the GRF were given by

\begin{equation}
\begin{aligned}
    \mu &= \frac{1}{2} - \frac{1}{\epsilon \sqrt{\pi}} \, \mathrm{erf}^{-1} \left( 1 - 2 \varphi \right) \exp \left( -\left[\mathrm{erf}^{-1} \left( 1 - 2 \varphi \right) \right]^2 \right) \\
    \textrm{and} \quad \varsigma^2 &= \frac{1}{2 \pi \epsilon^2} \exp \left( -2 \left[\mathrm{erf}^{-1} \left( 1 - 2 \varphi \right) \right]^2 \right),
\end{aligned}
\label{eqn:mean_and_variance}
\end{equation}\smallskip

\noindent where $\mathrm{erf}^{-1}( \cdot )$ is the inverse of the error function. $\mu$ is assumed to be constant with respect to $z_o$ and $\theta_o$. The insult propensity $\varphi$ corresponds to the fraction of $\vartheta^\ast$ values greater than 0.5, and $\epsilon$ corresponds to the slope of the cumulative distribution function of $\vartheta^\ast$ when $\vartheta^\ast = 0.5$. The covariance function, enforced to be periodic in $\theta$, was defined as 

\begin{equation}
\begin{aligned}
    \kappa(z_o, \theta_o, z_o', \theta_o') &= \varsigma^2 \exp \left( -\frac{1}{2} \left[ \left( \frac{D_\theta \left( \theta_o, \theta_o' \right)}{L_\theta} \right)^2 + \left( \frac{D_z \left( z_o, z_o' \right)}{L_z} \right)^2 \right] \right) \\
    \textrm{with} \quad D_\theta \left( \theta_o, \theta_o' \right) &= 2 r_o \sin{ \left( \frac{1}{2}\left| \theta_o - \theta_o' \right| \right) } \quad \text{and} \quad D_z \left( z_o, z_o' \right) = \left| z_o - z_o' \right|.
\end{aligned}
\end{equation}\smallskip

When mapping $\vartheta^\ast$ to the vessel wall nodes in the finite element simulations, a multivariate Gaussian distribution $\mathcal{N}$ with mean vector $\bm{\mu} = \mu \mathbf{1}$ and covariance matrix $\bm{\Sigma}$ was adopted, where $\Sigma_{ij} = \Sigma_{ji} = \kappa(z_{o,i}, \theta_{o,i}, z_{o,j}, \theta_{o,j})$. Partitioning the mesh into the set of interior nodes $a$ and the set of boundary nodes $b$, we conditioned the distribution of $\vartheta_a^\ast$ on the enforced value of $\vartheta_b^\ast$ using

\begin{equation}
\begin{aligned}
    \bm{\mu}_a' &= \bm{\mu}_a + \bm{\Sigma}_{ab} \bm{\Sigma}_{bb}^{-1} \left( \vartheta_b^\ast \bm{1} - \bm{\mu}_b \right) = \mu + \bm{\Sigma}_{ab} \bm{\Sigma}_{bb}^{-1} \left( \vartheta_b^\ast - \mu \right) \bm{1} \\
    \bm{\Sigma}_{aa}' &= \bm{\Sigma}_{aa} - \bm{\Sigma}_{ab} \bm{\Sigma}_{bb}^{-1} \bm{\Sigma}_{ba} \\
    \bm{\mu}_b' &= \vartheta_b^\ast \bm{1} \\
    \bm{\Sigma}_{ab}' &= \bm{0}, \quad \bm{\Sigma}_{ba}' = \bm{0}, \quad \bm{\Sigma}_{bb}' = \bm{0}.
\end{aligned}
\end{equation}\smallskip

\noindent After $\vartheta_i^\ast$ is sampled from $\mathcal{N} \left( \bm{\mu}', \bm{\Sigma}' \right)$, a cumulative distribution function (CDF)/inverse-CDF transformation was performed so that the overall distribution of $\vartheta^\ast$ values in each random instance of $\vartheta_i^\ast$ matches the desired $\mathcal{N} \left( \mu, \varsigma^2 \right)$. Specifically,

\begin{equation}
    \left( \vartheta_i^\ast \right)' = \Phi^{-1} \left( F \left( \vartheta_i^\ast \right); \mu, \varsigma^2 \right),
\end{equation}\smallskip

\noindent where $F$ is the CDF of the generated random field values and $\Phi^{-1}$ is the inverse CDF of the normal distribution with mean $\mu$ and variance $\varsigma^2$. Finally, the insult field values were censored using $\vartheta_i = \min \left( \max \left( \left( \vartheta_i^\ast \right)', 0 \right), 1 \right)$, and the resulting profiles corresponded in turn to patterns of compromised elastic fiber integrity and mechanosensing.

\section*{Finite element model}
The aortic geometry and material properties were based on biaxial mechanical and histological data from a common mouse model of Marfan syndrome ($Fbn\it1^{C1041G/+}$), estimated with previously discussed methods \cite{Latorre2018b}. The aorta was meshed with quadratic hexahedral elements with a radial, circumferential, and axial resolution of $1\times20\times20$, selected after previous mesh sensitivity studies to achieve sufficient accuracy while balancing computational efficiency. Using a custom plugin in the FEBio solver (\url{FEBio.org}), G\&R of the vessel in response to the combined elastic fiber integrity and mechanosensing insults was computed under constant systolic loading conditions of 120 mmHg over a series of gradual pseudo-time increments representing evolution over the course of weeks. After the final (systolic) geometry was achieved, the G\&R was arrested while the internal pressure was adjusted to a diastolic condition of 80 mmHg, a time frame corresponding to that of a cardiac cycle. Dilatation and distensibility quantities were evaluated on a nodal basis, that is, a two-dimensional domain of $41 \times 40$ nodes axially and circumferentially (noting periodicity in the circumferential direction). We performed circular padding in the circumferential direction (i.e., appended one additional column by repeating the first column) to produce input data with dimensions of $41 \times 41$.

\section*{Region-specific error evaluations}
In addition to computing prediction errors across the entire aortic domain, we also evaluated errors in the most affected regions of the vessel, namely any location with a normalized insult value of greater than 50\%. The overall errors are summarized in \autoref{tab:filtered-rel-l2}.

\begin{table}[!h]
    \centering
    \caption{Relative $\mathcal{L}_2$ errors for all network-input data combinations predicting combined insult contributors, evaluated over all testing cases. Errors are evaluated over both the total domain as well as only regions with $\geq$50\% normalized insult. The best results across all combinations (lowest $\mathcal{L}_2$ error) are highlighted by boldface in both approaches.}
    \resizebox{\textwidth}{!}{
    \def\arraystretch{1}
    \begin{tabular}{l >{\centering\arraybackslash}p{3.5cm} >{\centering\arraybackslash}p{3.5cm} >{\centering\arraybackslash}p{3.5cm} >{\centering\arraybackslash}p{3.5cm}}
    \hline
    & \multicolumn{4}{c}{Overall Relative $\mathcal{L}_2$ Error: Total Domain / $\geq$50\% Insult} \\
    & $d$ Grayscale & $d$ Heat & $d$ \& $\mathcal{D}$ Grayscale & $d$ \& $\mathcal{D}$ Heat \\
    \hline
    \bf{Eln Fiber Integrity} & & & & \\
    \quad CNN-DeepONet & 0.0614 / 0.0541 & 0.0534 / 0.0439 & 0.0244 / 0.0169 & 0.0237 / 0.0150 \\
    \quad FNN-DeepONet & 0.0689 / 0.0636 & 0.1275 / 0.1183 & 0.0329 / 0.0278 & 0.0276 / 0.0202 \\
    \quad UNet         & 0.0560 / 0.0528 & 0.0542 / 0.0515 & {\bf0.0176} / 0.0132 & 0.0182 / {\bf0.0130}\\
    \quad LNO          & 0.1708 / 0.1710 & 0.1584 / 0.1582 & 0.0263 / 0.0236 & 0.0223 / 0.0192 \\
    \hline
    \bf{Mechanosensing} & & & & \\
    \quad CNN-DeepONet & 0.0738 / 0.0655 & 0.0596 / 0.0481 & 0.0284 / 0.0213 & 0.0257 / 0.0168 \\
    \quad FNN-DeepONet & 0.0856 / 0.0754 & 0.1564 / 0.1360 & 0.0444 / 0.0403 & 0.0377 / 0.0310 \\
    \quad UNet         & 0.0675 / 0.0633 & 0.0628 / 0.0596 & 0.0187 / 0.0152 & {\bf0.0182} / {\bf0.0147} \\
    \quad LNO          & 0.1955 / 0.1952 & 0.1829 / 0.1819 & 0.0299 / 0.0269 & 0.0229 / 0.0200 \\
    \hline
    \end{tabular}}
    \label{tab:filtered-rel-l2}
    \end{table}

\begin{figure}[t]
\centering
\includegraphics[width=\textwidth]{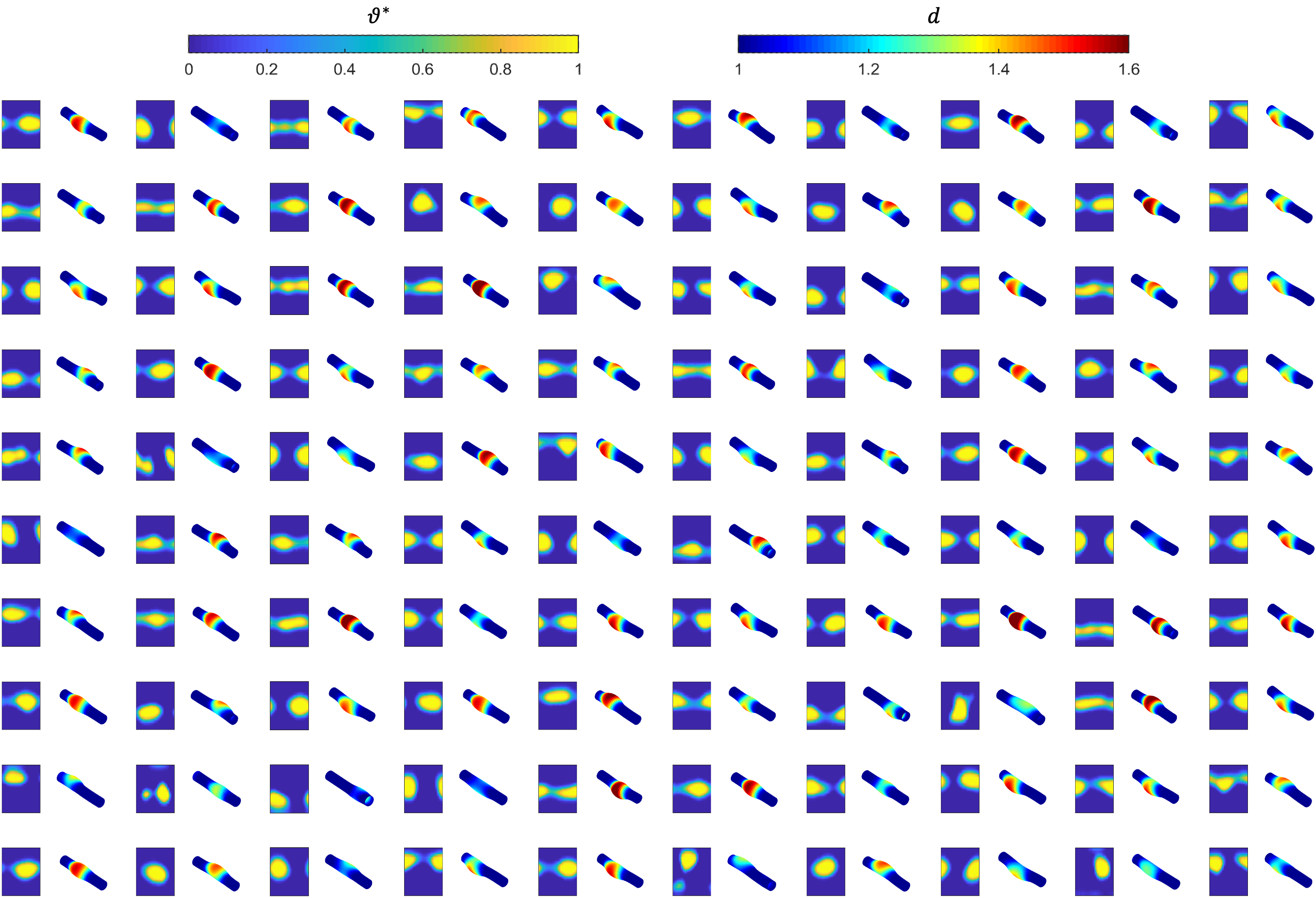}
\caption{{\bf Dilatation and distensibility maps.} Effects on dilatation $d$ and distensibility $\mathcal{D}$ depending on the dominating contributor (compromised elastic fiber integrity or dysfunctional mechanosensing) in the combined insult. (a) Normalized insult field $\vartheta^\ast$. (b) Degree of compromised integrity of elastic fibers ($\vartheta_{c^e} \in [0,0.48]$, decreasing left-to-right) and dysfunctional mechanosensing ($\vartheta_{\delta} \in [0,0.28]$, increasing left-to-right) with spatial distributions defined by the normalized insult profile. (c) Heat maps for dilatation and distensibility for each combined insult. (d) Normalized 8-bit grayscale maps.}
\label{fig:dildis-effects}
\end{figure}

\begin{figure}[t]
\centering
\includegraphics[width=\textwidth]{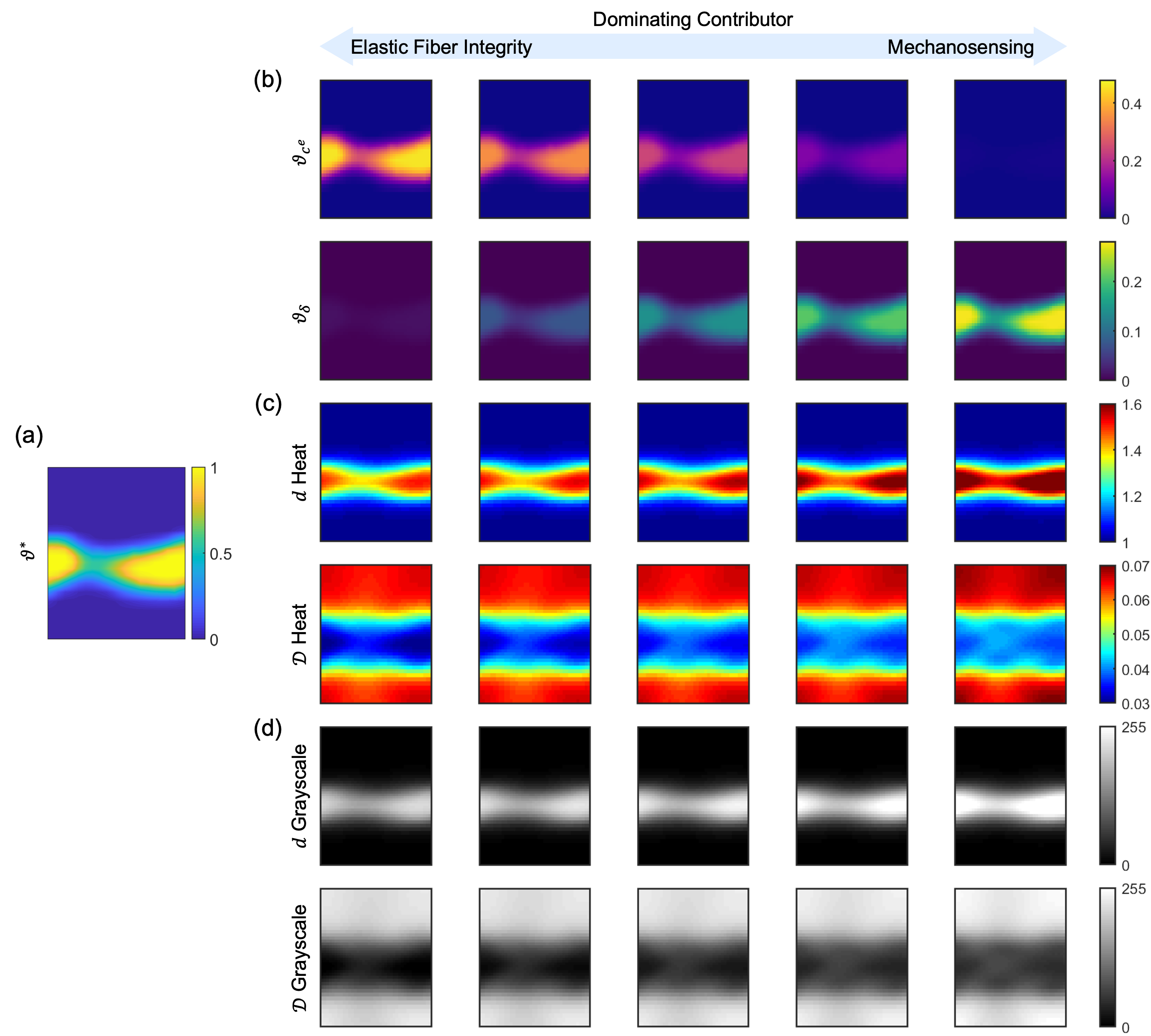}
\caption{{\bf Illustrative results.} Insult fields ($\vartheta^\ast$) and resulting TAA geometries colored by dilatation ($d$). A total of 100 ($10 \times 10$ panels) unique spatial distributions were generated with Gaussian random fields, and each profile was assigned a random combination of compromised elastic fiber integrity ($\vartheta_{c^e} \in [0,0.48]$) and dysfunctional mechanosensing ($\vartheta_{\delta} \in [0,0.28]$). Shown here is the dilatation corresponding to the most mechanosensing-dominated combination. We set the average circumferential and axial lengths to 4.5 mm and 4.5 mm, respectively, with a boundary softness of 0.2 and an overall insult area of roughly 23\% the total vessel area.}
\label{fig:all-taas}
\end{figure}

\begin{figure}[t]
\centering
\includegraphics[width=\textwidth]{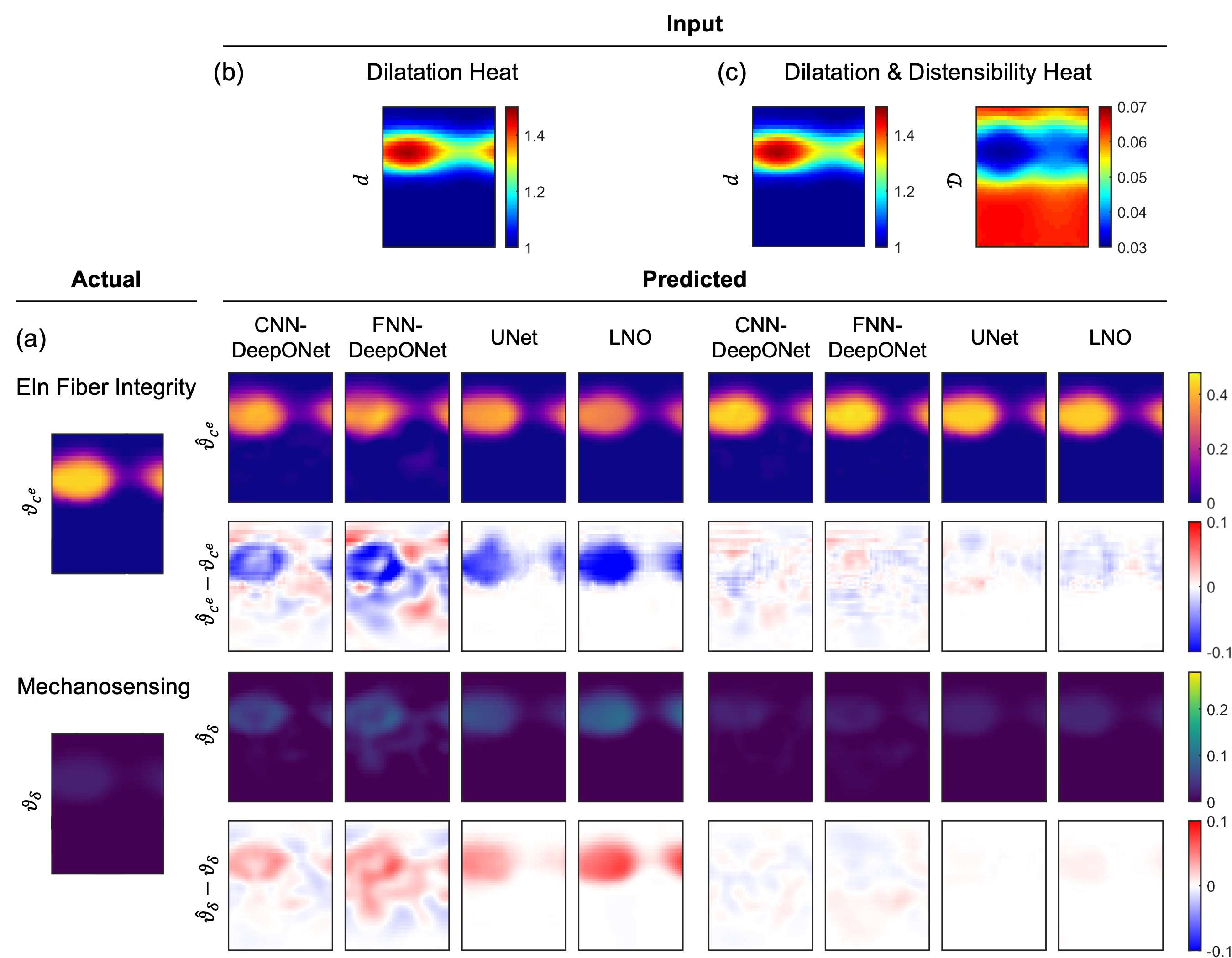}
\caption{{\bf Predictions from all four architectures for an elastic fiber integrity-dominated combined insult.} (a) The ground truth combined-insult field consisted of contributions of both compromised elastic fiber integrity ($\vartheta_{c^e}$) and dysfunctional mechanosensing ($\vartheta_{\delta}$) in the FE simulation to generate the TAA with dilatation and distensibility profiles shown in (b--c). Predictions ($\hat{\vartheta}_i$) and absolute errors ($\hat{\vartheta}_i - \vartheta_i$ ($i=c^e,\delta$)) are shown for CNN-DeepONet, FNN-DeepONet, UNet, and LNO trained on (b) dilatation heat maps only and (c) dilatation and distensibility heat maps.}
\label{fig:ef-allnets-heat}
\end{figure}

\begin{figure}[t]
\centering
\includegraphics[width=\textwidth]{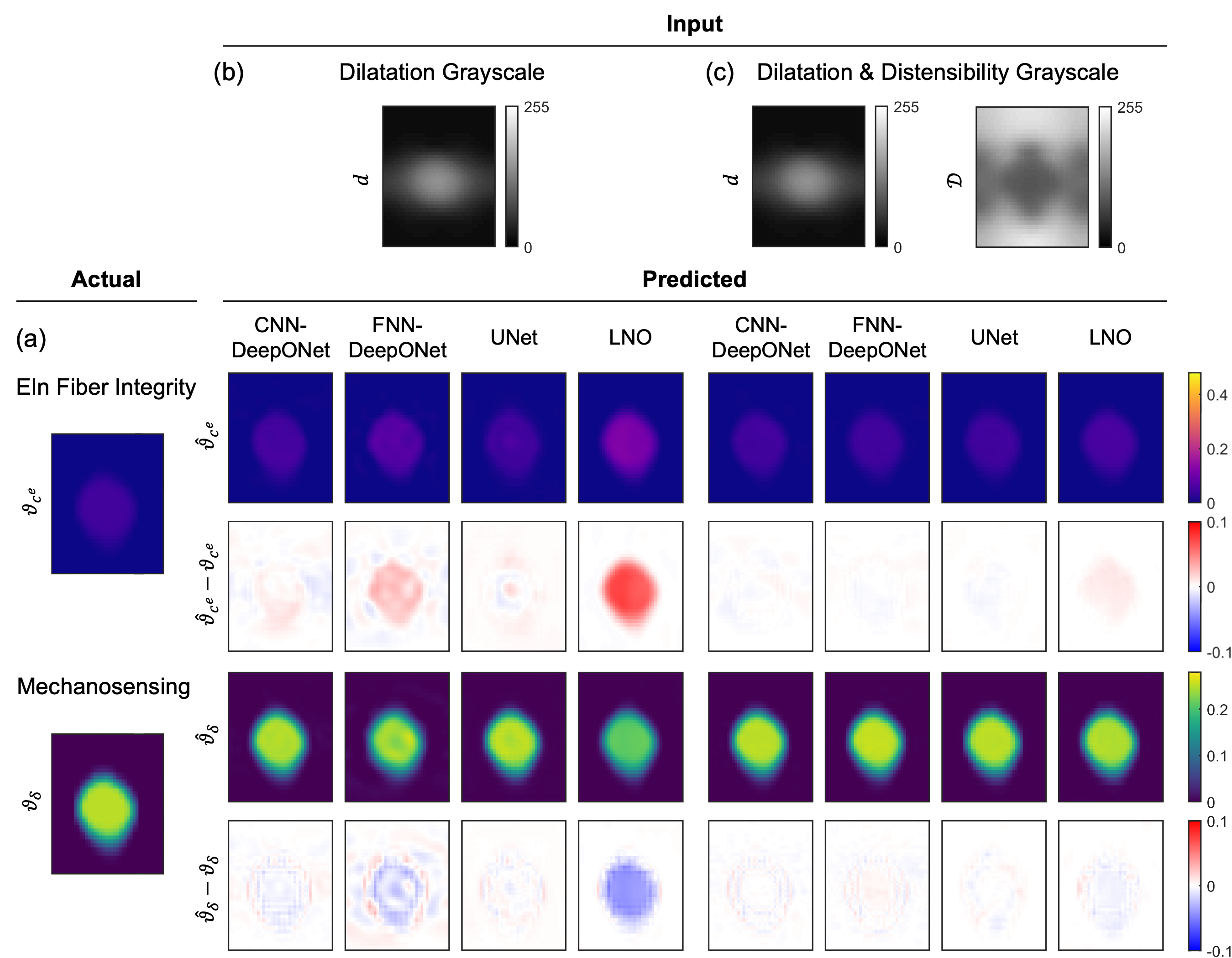}
\caption{{\bf Predictions from all four architectures for a mechanosensing-dominated combined insult.} Similar to \autoref{fig:ef-allnets-heat} except for grayscale input maps. (a) The ground truth combined-insult field consisted of contributions of both compromised elastic fiber integrity ($\vartheta_{c^e}$) and dysfunctional mechanosensing ($\vartheta_{\delta}$) superimposed in the FE simulation to generate the TAA with dilatation and distensibility profiles shown in (b--c). Predictions $\hat{\vartheta}_i$ and absolute errors $\hat{\vartheta}_i - \vartheta_i$ ($i=c^e,\delta$) are shown for CNN-DeepONet, FNN-DeepONet, UNet, and LNO trained on (b) dilatation grayscale maps only and (c) dilatation and distensibility grayscale maps.}
\label{fig:ms-allnets-gray}
\end{figure}

\begin{figure}[t]
\centering
\includegraphics[width=\textwidth]{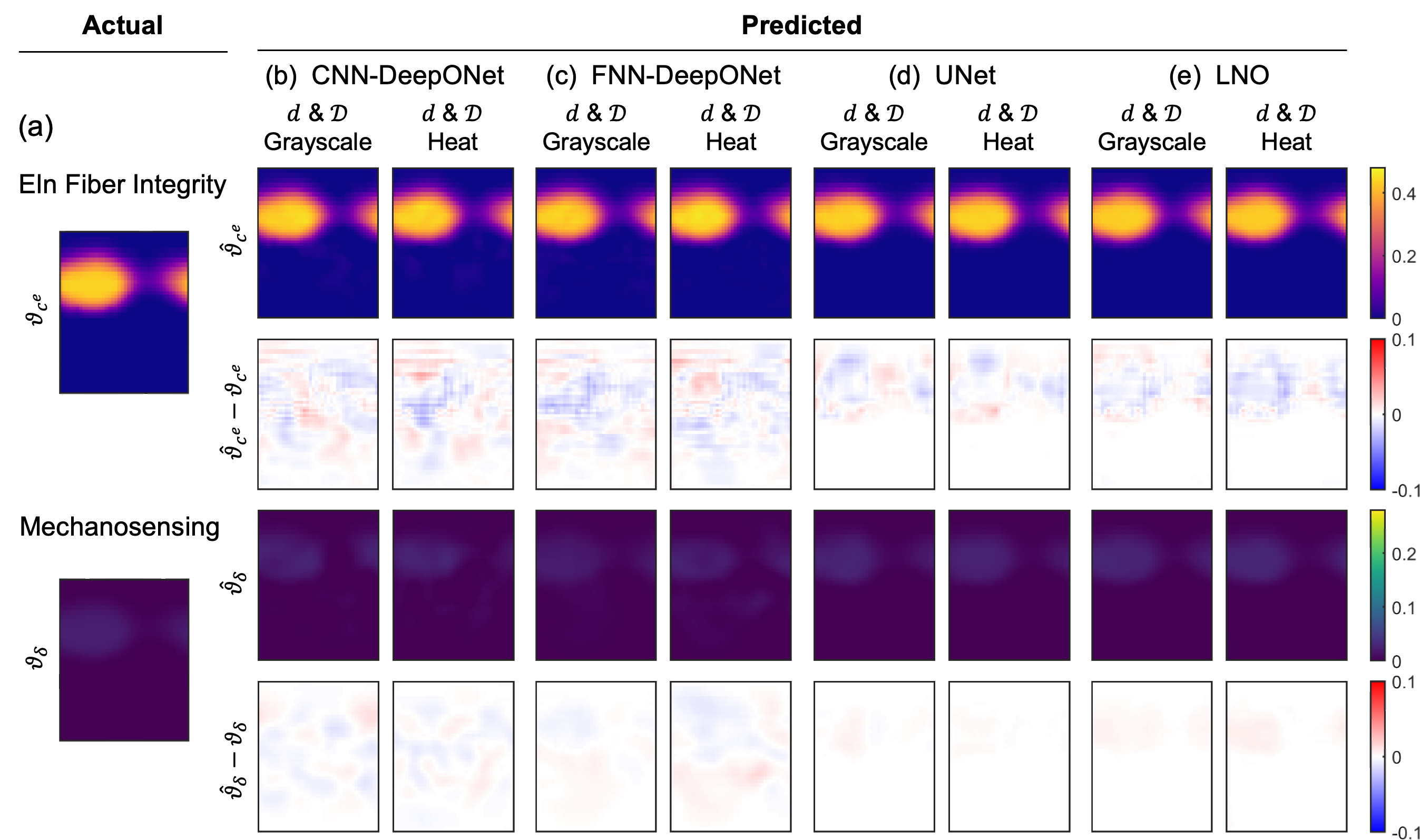}
\caption{{\bf Effects of grayscale and heat map data inputs.} All networks were trained on dilatation ($d$) and distensibility ($\mathcal{D}$) maps. (a) Ground truths for elastic fiber integrity and mechanosensing contributions. Predictions ($\hat{\vartheta}_i$) and computation of absolute errors ($\hat{\vartheta}_i - \vartheta_i$ ($i=c^e,\delta$)) are shown for (b) CNN-DeepONet, (c) FNN-DeepONet, (d) UNet, and (e) LNO.}
\label{fig:all-cases-dildis}
\end{figure}

\begin{figure}[t]
\centering
\includegraphics[width=\textwidth]{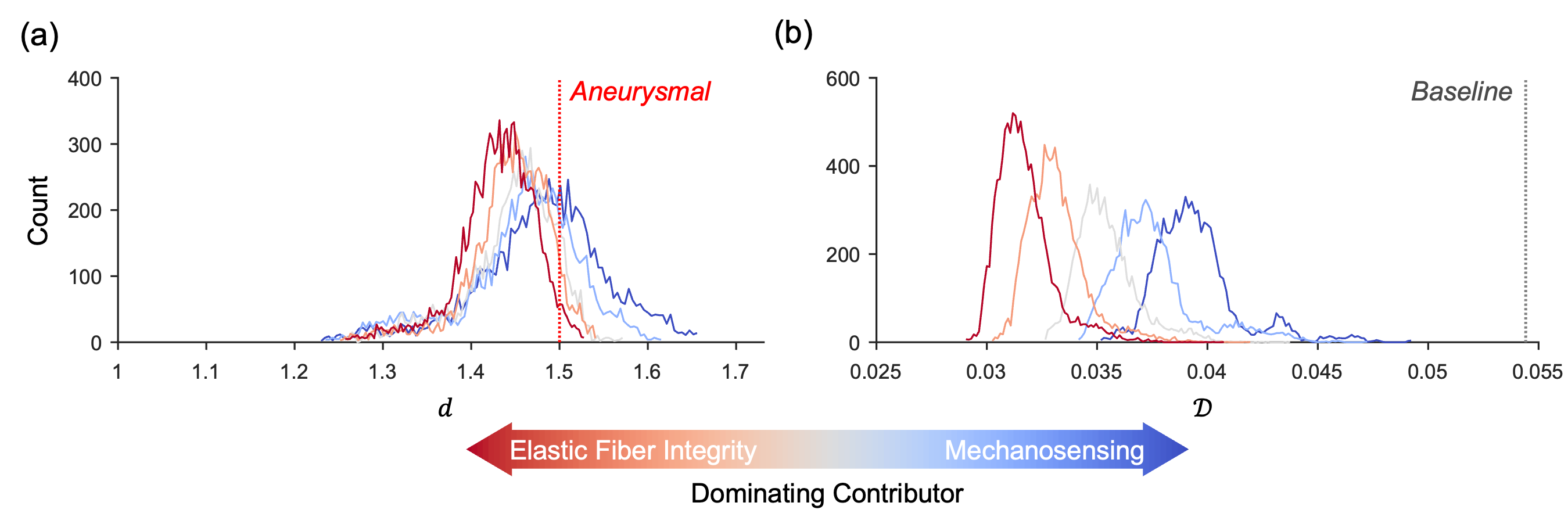}
\caption{{\bf Effects of insult contributor on dilatation and distensibility.} Histograms of all pairs of (a) maximum dilatation and (b) minimum distensibility over all points within the insult region in 500 synthetic TAAs. Points are colored based on dominating combined insult contributor (spectrum shown at bottom). Aneurysmal dilatations are indicated as $d > 1.5$. Baseline distensibility (no insult applied) is indicated by the dotted gray line.}
\label{fig:dil-vs-dis}
\end{figure}

\begin{figure}[t]
\centering
\includegraphics[width=\textwidth]{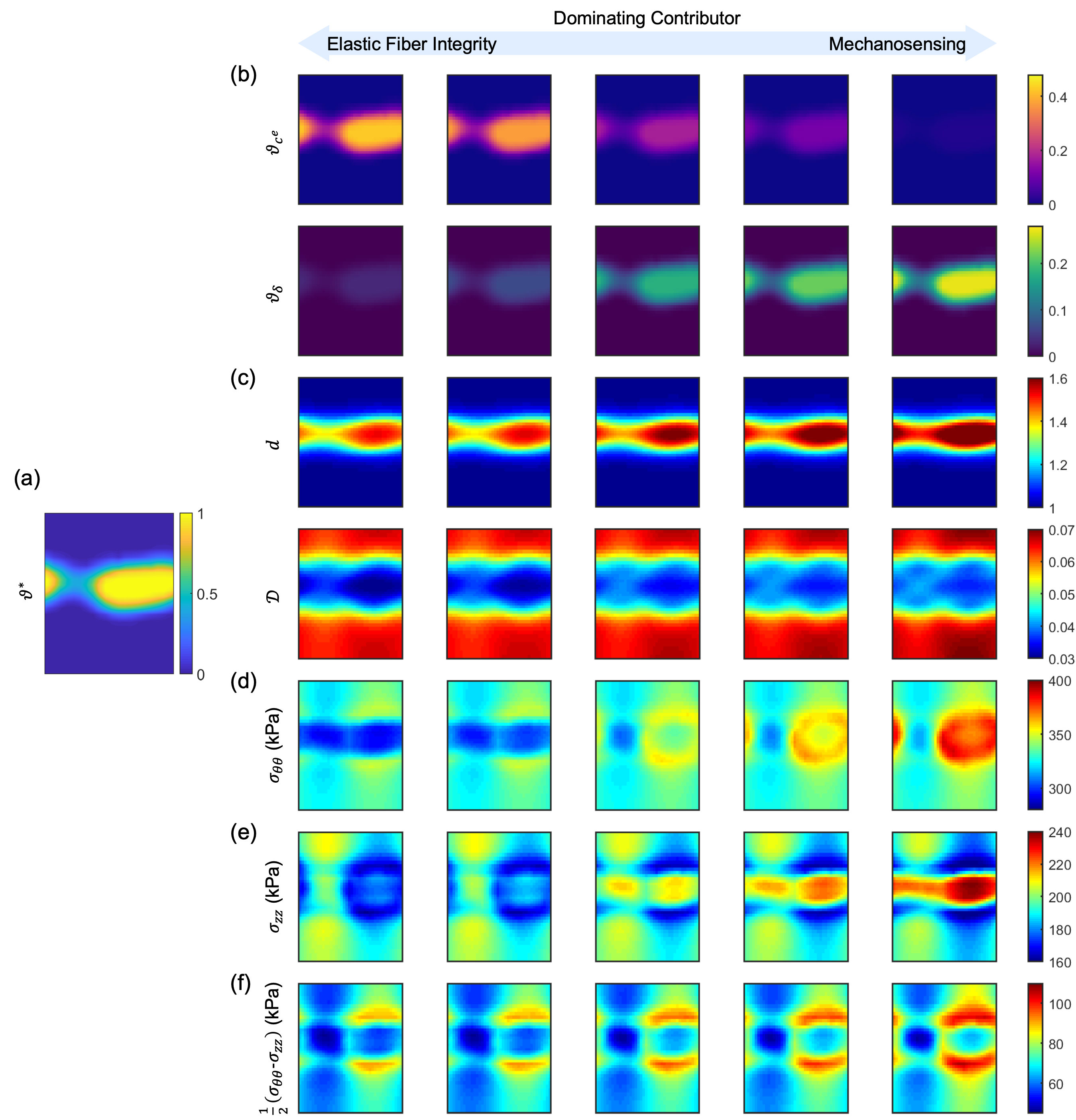}
\caption{{\bf Effects on geometry and mechanical properties.} Similar to \autoref{fig:dildis-effects} but with a different insult profile. (a) Normalized insult field ($\vartheta^\ast$). (b) Degree of compromised integrity of elastic fibers ($\vartheta_{c^e} \in [0,0.48]$, decreasing left-to-right) and dysfunctional mechanosensing ($\vartheta_{\delta} \in [0,0.28]$, increasing left-to-right). (c) Heat maps for dilatation and distensibility for each combined insult. (d) Circumferential and (e) axial components of the Cauchy stress. (f) Estimated intramural shear stress computed as $(\sigma_{\theta\theta} - \sigma_{zz})/2$. Stress estimations underscore that the mechanobiological mechanisms driving dilatation may have a profound impact on the mechanical stability of the aneurysm and yet may appear nearly identical based on geometry alone.}
\label{fig:dildisstress-effects}
\end{figure}

\end{document}